%% file: klcf.tex
\documentclass{article}
\usepackage[preprint]{klcf}

\usepackage[utf8]{inputenc} % allow utf-8 input
\usepackage[T1]{fontenc}    % use 8-bit T1 fonts
\usepackage{hyperref}       % hyperlinks
\usepackage{url}            % simple URL typesetting
\usepackage{booktabs}       % professional-quality tables
\usepackage{amsfonts}       % blackboard math symbols
\usepackage{nicefrac}       % compact symbols for 1/2, etc.
\usepackage{microtype}      % microtypography
\usepackage{xcolor}         % colors

\usepackage{titletoc}

\usepackage[most]{tcolorbox}
\tcbuselibrary{breakable, skins, theorems}

\usepackage{lipsum} % for dummy text
\usepackage{multirow}
\usepackage{makecell}
\usepackage[table]{xcolor}
\usepackage{amssymb}  % \checkmark
\usepackage{bbding}   % \Checkmark, \XSolid
\usepackage{pifont}   %
\usepackage{wasysym}  % \checked, \XBox
\usepackage{fontawesome5}
\usepackage{academicons}
\usepackage{enumitem}
\usepackage{tabularx}
\usepackage{ragged2e} % \RaggedRight
\usepackage{caption}
\usepackage{array}
\usepackage{adjustbox}

\usepackage{natbib}
\setcitestyle{numbers,square}
\input{math_commands.tex}

\newcounter{promptcounter}

\makeatletter
\renewcommand{\@fnsymbol}[1]{%
  \ifcase#1\or \dagger \or \ddagger \or \S \or \P \or \|\or **\fi
}
\makeatother

\usepackage{graphicx}
\usepackage{xspace}
\setcitestyle{square,comma}

\newcommand{\huggingface}{\raisebox{-1.5pt}{\includegraphics[height=1.05em]{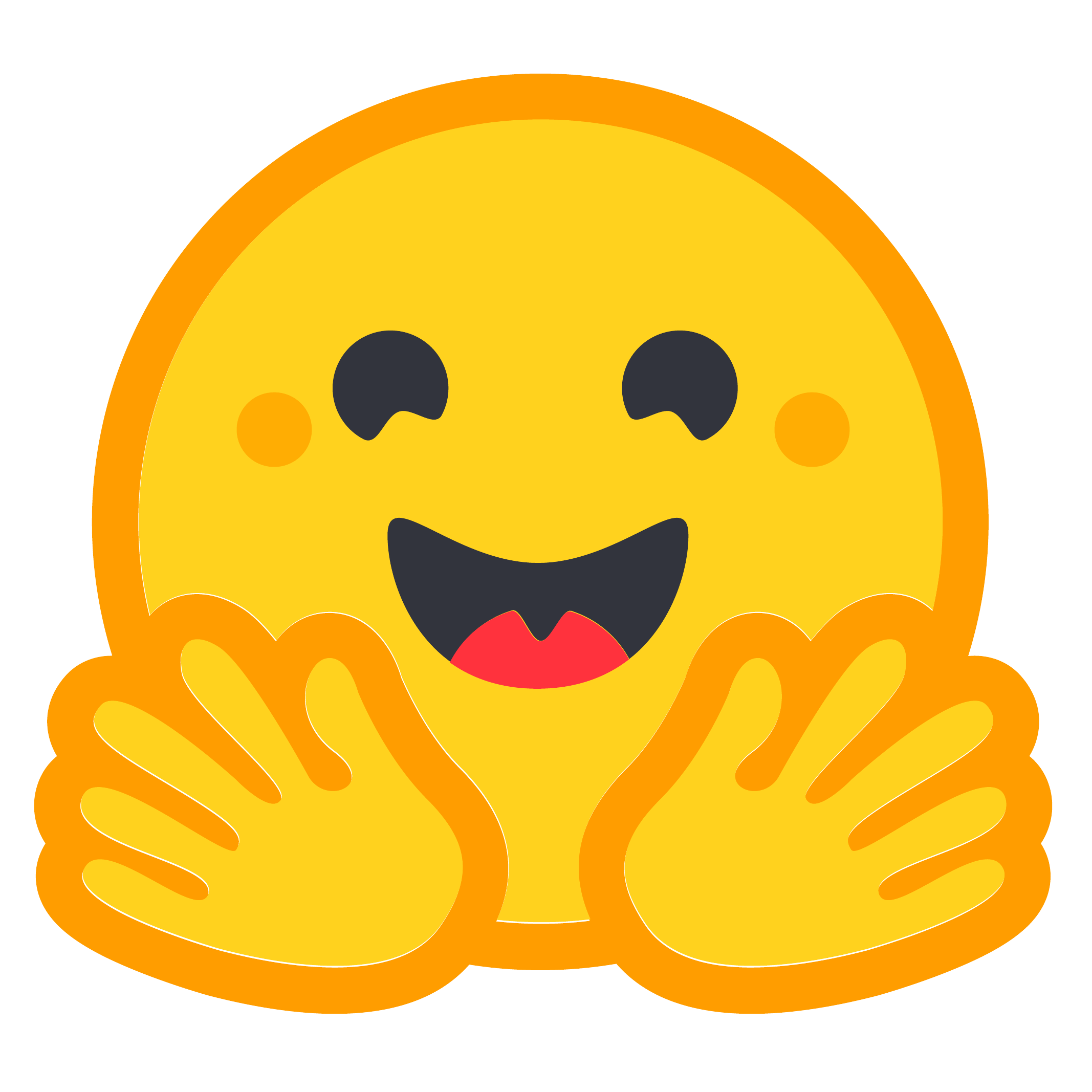}}\xspace}

\newcommand{\github}{\raisebox{-1.5pt}{\includegraphics[height=1.05em]{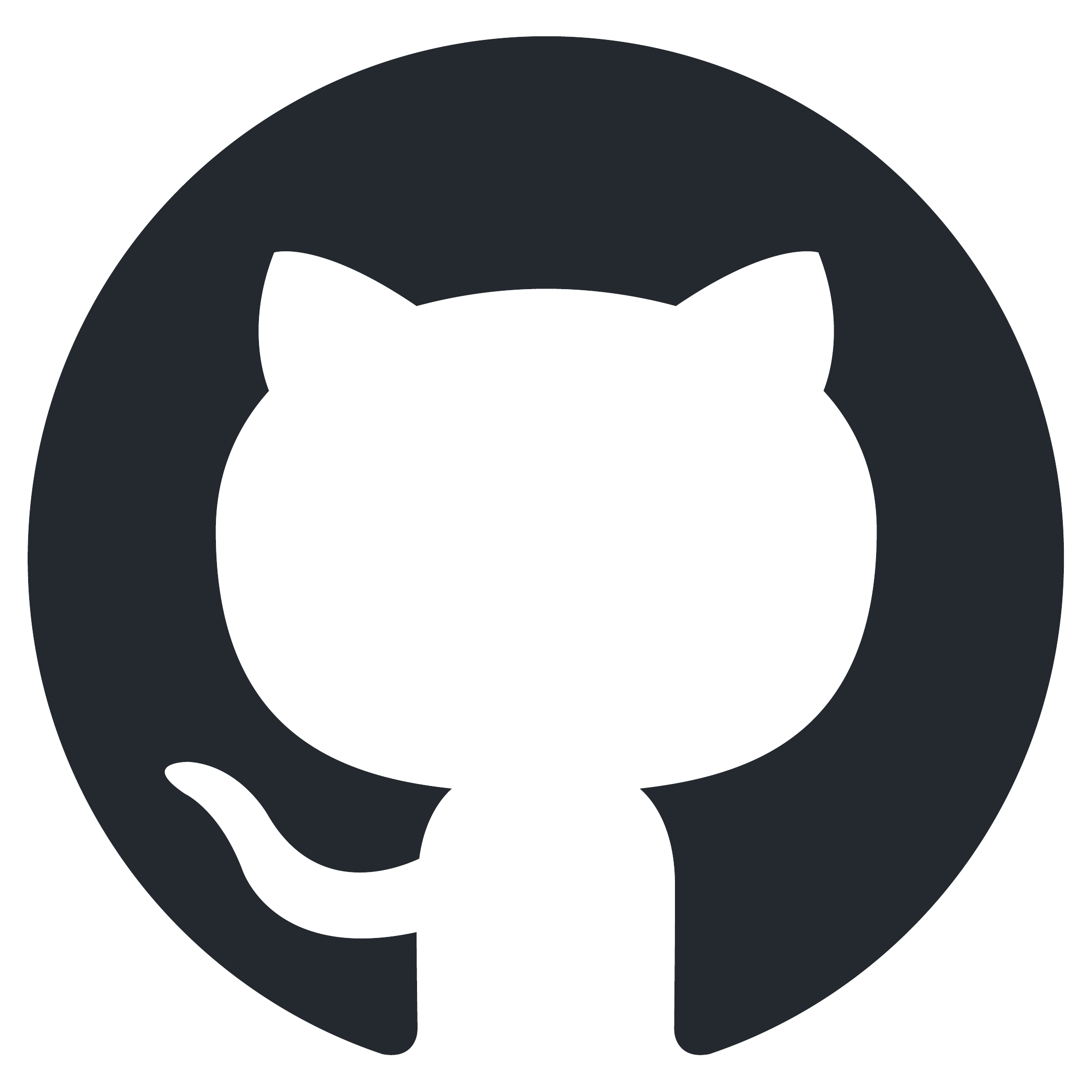}}\xspace}

\title{Knowledge-Level Consistency Reinforcement Learning: Dual-Fact Alignment for Long-Form Factuality}

\author{Junliang Li, Yucheng Wang, Yan Chen, Yu Ran \\
\textbf{Ruiqing Zhang, Jing Liu\thanks{
  Corresponding Author. Email: liujing46@baidu.com.}, Hua Wu, Haifeng Wang} \\
Baidu Inc. \\
\texttt{\{lijunliang03,wangyucheng01,chenyan22,ranyu03\}@baidu.com}\\
\texttt{\{zhangruiqing01,liujing46,wu\_hua,wanghaifeng\}@baidu.com}\\
}

\begin{document}

\maketitle

\input{abstract}

\input{introduction}  %

\input{method}

\input{experiment}  %

\input{related_work}

\input{conclusion}  %

\newpage
% \section*{References}
\bibliography{klcf}
\bibliographystyle{ieee}

%%%%%%%%%%%%%%%%%%%%%%%%%%%%%%%%%%%%%%%%%%%%%%%%%%%%%%%%%%%%

\newpage
\appendix
% \startcontents[appendices]     
% \printcontents[appendices]{}{0}{\setcounter{tocdepth}{2}}
% \renewcommand{\contentsname}{Appendices}

\startcontents[appendices]
\renewcommand{\contentsname}{Appendices}
\printcontents[appendices]{}{0}{\setcounter{tocdepth}{2}}

\input{appendix}

\end{document}

%% file: math_commands.tex
\def\eqref#1{equation~\ref{#1}}
\def\1{\bm{1}}

\DeclareMathAlphabet{\mathsfit}{\encodingdefault}{\sfdefault}{m}{sl}
\SetMathAlphabet{\mathsfit}{bold}{\encodingdefault}{\sfdefault}{bx}{n}

%% file: abstract.tex
\begin{abstract}
Hallucination in large language models (LLMs) during long-form generation remains difficult to address under existing reinforcement learning from human feedback (RLHF) frameworks, as their preference rewards often overlook the model's own knowledge boundaries. In this paper, we propose the \textbf{K}nowledge-\textbf{L}evel \textbf{C}onsistency Reinforcement Learning \textbf{F}ramework (\textbf{KLCF}), which re-examines this problem from a distribution alignment perspective. KLCF formalizes long-form factuality as a bidirectional distribution matching objective between the policy model's expressed knowledge distribution and the base model's parametric knowledge distribution: under the constraint that generation must not exceed the support set of the base knowledge, the objective maximizes coverage of high-probability facts, thereby jointly optimizing precision and recall. To achieve this, we design a Dual-Fact Alignment mechanism that approximates the recall term using a factual checklist constructed by sampling from the base model, and constrains hallucinations with a lightweight truthfulness reward model. Both components are jointly optimized and require no external retrieval throughout training. Experimental results demonstrate that KLCF consistently improves factuality metrics across multiple long-form benchmarks and model scales, effectively alleviating hallucination and over-conservatism while maintaining efficiency and scalability.

\github \href{https://github.com/ki-ljl/KLCF}{\textcolor{blue}{\texttt{https://github.com/ki-ljl/KLCF}}} \\
\huggingface\href{https://huggingface.co/JunliangLi/models}{\textcolor{blue}{\texttt{https://huggingface.co/collections/JunliangLi/klcf}}}
\end{abstract}

%% file: introduction.tex
\section{Introduction}
LLMs excel at tasks such as question answering, summarization, and complex reasoning~\citep{li2024flexkbqa,xu2024generate,zhang2024comprehensive,gupta2025autosumm,wei2022chain,shao2024deepseekmath,zhang2025ratt}, yet hallucinations—generating content that conflicts with established facts—remain a critical threat to reliability \citep{huang2025survey,wang2024factuality}, especially in long‑form generation where early errors can snowball \citep{zhang2023language,ji2023survey,huang2025survey}. Moreover, existing RLHF frameworks~\citep{bai2022training} rely on preference‑based rewards but overlook the model’s parametric knowledge boundary, inadvertently exacerbating the so‑called ``hallucination tax''~\citep{cotra2021ai,sharma2023towards}.

To evaluate and address the factuality and hallucination issues in long-form generation, a commonly employed approach is to decompose the model's output into atomic facts and verify them through external retrieval. The early work FActScore~\citep{min2023factscore} and FacTool~\citep{chern2023factool} established this paradigm. Subsequently, methods like SAFE~\citep{wei2024long} and VeriScore~\citep{song2024veriscore} further refined the fact extraction and verification pipeline. Methods like FactTune-FS~\citep{tian2023fine}, FLAME~\citep{lin2024flame} and FactAlign~\citep{huang2024factalign} use evaluators like FActScore to provide factuality alignment signals through external retrieval, successfully reducing hallucinations to some extent. However, most of these approaches are still confined to offline Reinforcement Learning (RL) scenarios. The low efficiency of external retrieval-based verification makes these methods difficult to scale for large-scale online RL applications. In terms of effectiveness, the main limitations of these approaches include a narrow focus on factual precision without considering factual recall, often leading models to favor conservative responses.
\begin{figure}
    \centering
	\includegraphics[width=1.0\textwidth]{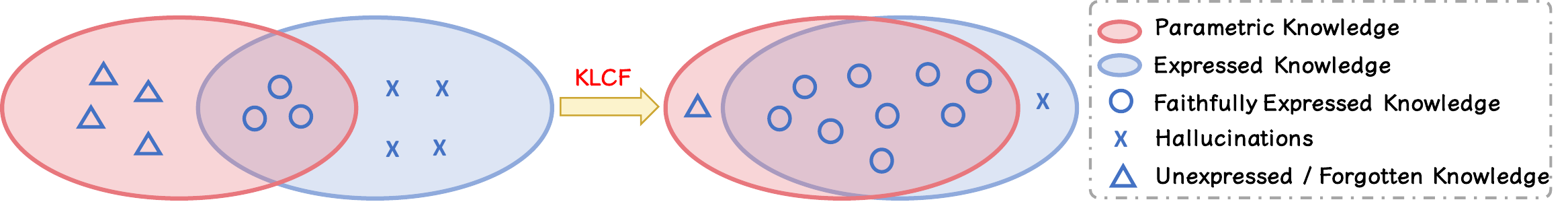}
		\caption{Conceptual illustration of KLCF. Conventional methods reduce hallucinations by shrinking Expressed Knowledge (hurting coverage), while KLCF expands the faithful overlap between Expressed Knowledge and Parametric Knowledge.}
	\label{fig:motivation}
\end{figure}
As illustrated in Fig.~\ref{fig:motivation}, the faithful overlap between the base model’s parametric knowledge and the policy’s expressed knowledge is not directly observable. Based on the principle that LLMs' parametric knowledge originates from pre-training~\citep{petroni-etal-2019-language,yu2024reveal}, we propose that the core objective of long-form factuality alignment is to increase the \textbf{knowledge-level consistency} between the aligned model's ``expressed knowledge'' and the base model's ``parametric knowledge''. From a statistical perspective, given a query $q$, the base model defines an implicit atomic fact distribution $P_{\mathrm{base}}(f|q)$, characterizing its parametric knowledge boundary; the policy model in turn induces its own expressed knowledge distribution $P_{\pi}(f|q)$. Long-form factuality alignment can thus be formalized as a constrained distribution matching problem: maximize the coverage of high-probability facts under $P_{\mathrm{base}}$ (high recall) while strictly constraining generation to facts within the base model’s knowledge boundary (high precision). Prior work~\citep{kadavath2022language,zhang-etal-2024-self} has shown that this objective can be achieved using only internal knowledge for reward signals.

Building on the above theoretical motivation, we propose the \textbf{K}nowledge-\textbf{L}evel \textbf{C}onsistency Reinforcement Learning \textbf{F}ramework (\textbf{KLCF}). KLCF approximates the distribution alignment objective through a dual-fact alignment mechanism: offline, it uses external retrieval to build a verifiable factual checklist from the base model’s generations and trains a lightweight truthfulness reward model, thereby eliminating reliance on external retrieval during online RL – the checklist approximates the recall term via Monte Carlo estimation, while the truthfulness reward model approximates the support constraint in the precision term. The two signals jointly optimize the precision and recall of expressed knowledge. Notably, although the reward signals are derived entirely from the base model’s internal knowledge, the resulting model achieves consistent gains on external verification benchmarks such as FActScore and VeriScore, confirming that internal-knowledge optimization serves as a reliable proxy for external factuality evaluation. The overall reward design is lightweight, retrieval-free during training, and readily scalable to large-scale online RL. Our main contributions are as follows:
\begin{enumerate}[leftmargin=*, label=(\arabic*)]
    \item We reframe long‑form factuality alignment as distribution matching that enforces consistency between the policy’s expressed knowledge and the base model’s parametric knowledge.
    \item We introduce a Dual‑Fact Alignment mechanism that jointly optimizes factual recall and precision through effective approximations of the alignment objective, forming a lightweight, retrieval‑free reward system for scalable online RL.
    \item Extensive experiments across benchmarks, scales, and reasoning modes show consistent gains.
\end{enumerate}

%% file: method.tex
\section{Knowledge-Level Consistency Reinforcement Learning}

\begin{figure}
    \centering
	\includegraphics[width=1.0\textwidth]{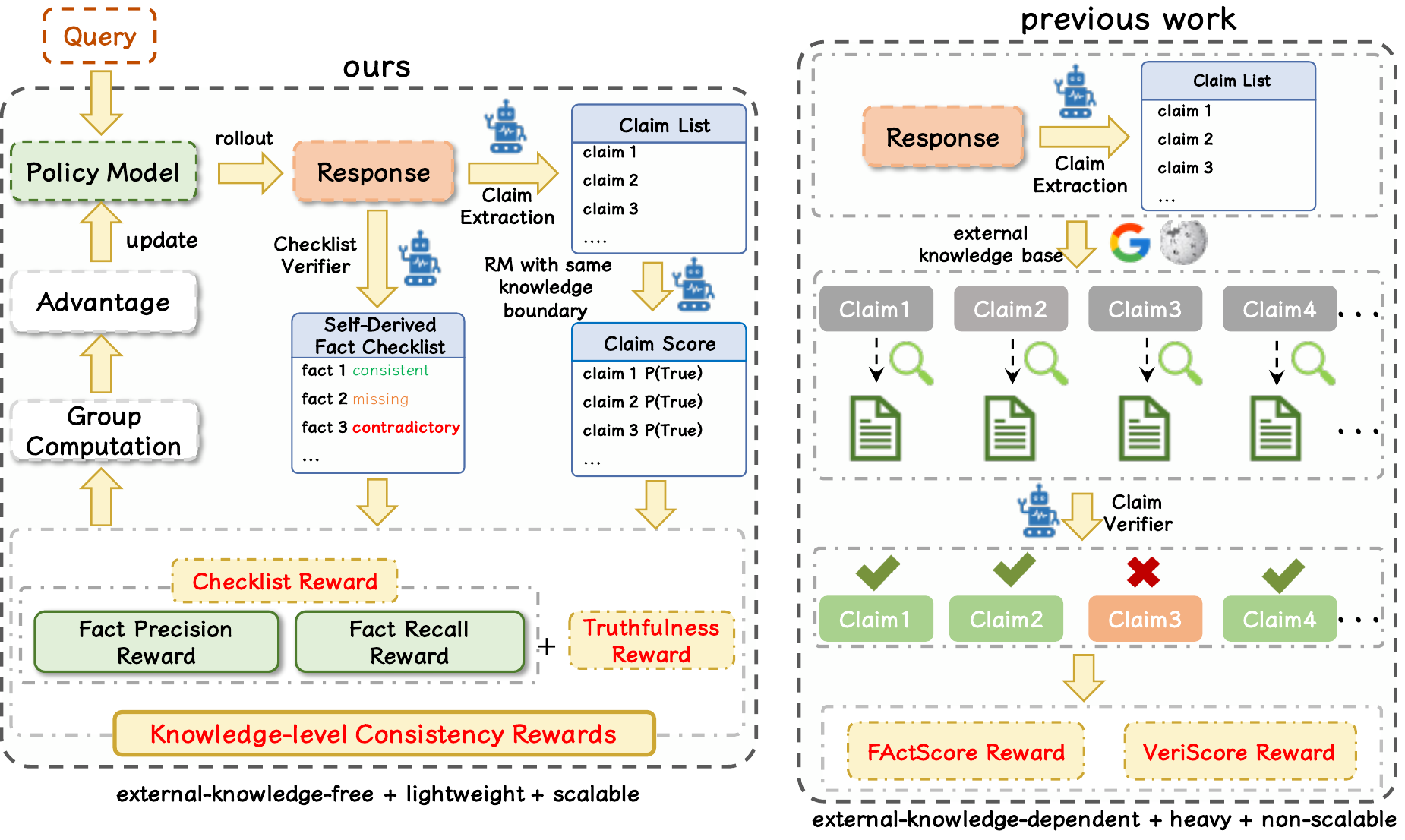}
		\caption{KLCF framework (Left) vs. Previous work (Right). KLCF achieves dual‑fact alignment via knowledge‑level consistency rewards computed offline with a factual checklist and a truthfulness reward model, eliminating costly online retrieval and enabling scalable RL.}
	\label{fig:framework}
\end{figure}
% Replacement text from \subsection{Problem Formulation} onward.

\subsection{Problem Formulation}\label{sec:problem_def}
We formulate long-form factual generation as a latent \emph{helpful-and-honest} objective. Given a query $q$, the policy model $\pi_\theta$ produces a response $y \sim \pi_\theta(\cdot \mid q)$, which is decomposed into a set of atomic claims $C(y)=\{c_1,c_2,\ldots\}$ by a claim extractor. Let $F^*(q)$ denote the latent set of all query-relevant and correct atomic facts that should ideally appear in a high-quality answer. We characterize factual helpfulness by the coverage of $F^*(q)$ and factual honesty by the truthfulness of the generated claims:
\begin{equation}
\mathrm{Recall}^*(q,y) = \frac{1}{|F^*(q)|} \sum_{f \in F^*(q)} \mathbb{I}[f \in C(y)],
\end{equation}
\begin{equation}
\mathrm{Precision}^*(q,y) = \frac{1}{|C(y)|} \sum_{c \in C(y)} \mathbb{I}[\mathrm{True}(c)].
\end{equation}
The corresponding ideal utility is
\begin{equation}\label{eq:ideal_obj}
U^*(q,y) = \alpha \cdot \mathrm{Recall}^*(q,y) + \beta \cdot \mathrm{Precision}^*(q,y) + \gamma \cdot H(q,y).
\end{equation}
where $H(q,y)$ denotes the holistic quality of the response, including factors such as instruction following, coherence, and readability.

This formulation also clarifies the role of knowledge consistency. Let $K_{\mathrm{para}}(q)$ denote the model's latent parametric knowledge and $K_{\mathrm{expr}}(q)$ its expressed knowledge under a fixed prompting and sampling setup. Because $K_{\mathrm{para}}(q)$ is not directly observable, the overlap $K_{\mathrm{para}}(q)\cap K_{\mathrm{expr}}(q)$ cannot be accessed exactly during online RL. In practice, we can only observe the claims that the model actually expresses, and then estimate which part of that expressed knowledge is both correct and reliable.

Accordingly, we separate \emph{evaluation} from \emph{training}. Final evaluation is externally grounded and model-agnostic: we measure query-relevant factual recall, claim support rate under retrieval-based verification, and judge win rate for overall answer quality. Training, in contrast, must remain lightweight enough for online RL. We therefore optimize a retrieval-free surrogate reward anchored to the base model's \emph{verified faithfully expressed knowledge}. This surrogate is not identical to the final evaluation metric, but it is designed to be operationally aligned with it while remaining practical for large-scale RL.

\subsection{Knowledge-Level Consistency Objective}\label{sec:klc_obj}
Our core idea is to increase the faithful expression of knowledge that the base model can already express reliably, while suppressing claims that likely fall outside this reliable region. For each query $q$, we sample the base model multiple times under a fixed prompting protocol and extract its claims. Since the expressed claims can contain both correct and incorrect statements, we retain only those that are externally verified as supported facts. This yields the set of \emph{verified faithfully expressed knowledge}:
\begin{equation}
K_{\mathrm{vfe}}(q) = \{f : f \in K^{\mathrm{base}}_{\mathrm{expr}}(q), \ \mathrm{Verify}(f)=\textsc{Support}\}.
\end{equation}
Operationally, $K_{\mathrm{vfe}}(q)$ is represented by a curated factual checklist $\Lambda(q)$ and by a truthfulness reward model trained on verified base-generated claims. Motivated by prior evidence that RL mainly elicits knowledge already acquired during pretraining, we use $K_{\mathrm{vfe}}(q)$ as an operational approximation to the base model's reliable knowledge boundary.

\subsubsection{Base-Anchored Surrogate Objective}\label{sec:alignment_obj}
Let $Q_{\mathrm{vfe}}(\cdot \mid q)$ denote the empirical distribution induced by the verified facts in $\Lambda(q)$, and let $s_{\mathrm{base}}(c)\in[0,1]$ denote the truthfulness score assigned to claim $c$ by the base-anchored reward model. We optimize the following surrogate objective:
\begin{equation}\label{eq:main_dual_obj}
\max_{\pi_\theta} \;
\underbrace{\mathbb{E}_{f \sim Q_{\mathrm{vfe}}(\cdot \mid q)} \left[ \Pr_{y \sim \pi_\theta(\cdot \mid q)}(f \in C(y)) \right]}_{\text{verified base knowledge recall}}
+
\lambda \underbrace{\mathbb{E}_{y \sim \pi_\theta(\cdot \mid q)} \left[ \frac{1}{|C(y)|+\delta} \sum_{c \in C(y)} s_{\mathrm{base}}(c) \right]}_{\text{base-anchored truthfulness}},
\end{equation}
where $\lambda > 0$ balances the two terms and $\delta > 0$ avoids division by zero. The first term encourages the policy to cover more verified facts that the base model can faithfully express, thereby improving factual helpfulness. The second term suppresses unfaithful expression by rewarding claims that are judged likely to lie inside the base model's reliable knowledge boundary.

Importantly, Eq.~(\ref{eq:main_dual_obj}) is a retrieval-free training surrogate rather than a direct estimator of the ideal objective in Eq.~(\ref{eq:ideal_obj}). Its purpose is to improve the faithful expression rate of relevant parametric knowledge using only signals that are available online during RL. We later validate this modeling choice empirically by showing that optimizing the base-anchored surrogate consistently improves externally grounded factual evaluation.

\subsubsection{Practical Reward Instantiation}\label{sec:monte_carlo}
Eq.~(\ref{eq:main_dual_obj}) is defined at the expectation level. During online RL, we instantiate it with single-sample rewards that can be computed from one generated response and the pre-built factual resources.

\textbf{Checklist-scope recall.} Given a response $y$ and checklist $\Lambda(q)$, each checklist item is compared against the claims in $y$ and labeled as \textsc{Consistent}, \textsc{Contradictory}, or \textsc{Missing}. Let $N_{\text{consistent}}$, $N_{\text{contradictory}}$, and $N_{\text{missing}}$ denote the corresponding counts. We define the \textbf{Fact Recall Reward} as
\begin{equation}\label{eq:recall}
R_{\text{recall}}(y) =
\frac{N_{\text{consistent}}}
{N_{\text{consistent}} + N_{\text{contradictory}} + N_{\text{missing}}}.
\end{equation}
This reward measures how much of the verified checklist is faithfully covered by the response, and serves as a practical proxy for the verified base knowledge recall term in Eq.~(\ref{eq:main_dual_obj}).

\textbf{Checklist-scope precision.} To sharpen supervision within the checklist scope, we additionally define the \textbf{Fact Precision Reward}
\begin{equation}\label{eq:precision}
R_{\text{precision}}(y) =
\begin{cases}
\dfrac{N_{\text{consistent}}}{N_{\text{consistent}} + N_{\text{contradictory}}}, & \text{if } N_{\text{consistent}} + N_{\text{contradictory}} > 0, \\
0, & \text{otherwise}.
\end{cases}
\end{equation}
This reward penalizes incorrect statements when the model attempts to cover checklist facts, and complements $R_{\text{recall}}$ by discouraging inaccurate coverage.

\textbf{Global truthfulness.} Since the checklist only supervises facts that have already been mined from the base model, it cannot directly constrain all content generated beyond the checklist scope. To address this, we train a lightweight \textbf{Truthfulness Reward Model} on verified base-generated claims (training details in Section~\ref{sec:verifier_training}). For each atomic claim $c \in C(y)$, the model outputs a score $s_{\mathrm{base}}(c)\in[0,1]$, where higher scores indicate that the claim is more likely to fall inside the base model's reliable knowledge boundary. We define the \textbf{Truthfulness Reward} as
\begin{equation}\label{eq:truth}
R_{\text{truth}}(y) =
\begin{cases}
\dfrac{1}{|C(y)|} \sum_{c \in C(y)} s_{\mathrm{base}}(c), & \text{if } |C(y)| > 0, \\
0, & \text{otherwise}.
\end{cases}
\end{equation}
\textbf{Checklist-aware truthfulness variant.} We also consider a lower-noise variant that only scores claims not already covered by the verified checklist. Specifically, all items in $\Lambda(q)$ are concatenated into a pseudo-response, and the claims in $C(y)$ are verified against this pseudo-response using the same protocol as the checklist reward. Let $C_M(y)$ denote the subset of claims marked as \textsc{Missing}. The variant reward is
\begin{equation}\label{eq:truth_var}
R_{\text{truth}}^{\text{variant}}(y) =
\begin{cases}
1, & \text{if } |C_M(y)| = 0, \\
\dfrac{1}{|C_M(y)|} \sum_{c \in C_M(y)} s_{\mathrm{base}}(c), & \text{otherwise}.
\end{cases}
\end{equation}
\textbf{Discussion.} The rewards above form our knowledge-level consistency (KLC) reward family. $R_{\text{recall}}$ increases the recall of verified base knowledge, $R_{\text{precision}}$ improves factual accuracy within the checklist scope, and $R_{\text{truth}}$ or $R_{\text{truth}}^{\text{variant}}$ provides a global retrieval-free signal against unfaithful expression. Together, they approximate the two desiderata in Eq.~(\ref{eq:main_dual_obj}) with per-sample signals that are efficient to compute during online RL.

\subsubsection{Data Preparation for Factual Alignment}\label{sec:rl_data}
\begin{figure}[t]
    \centering
	\includegraphics[width=1.0\textwidth]{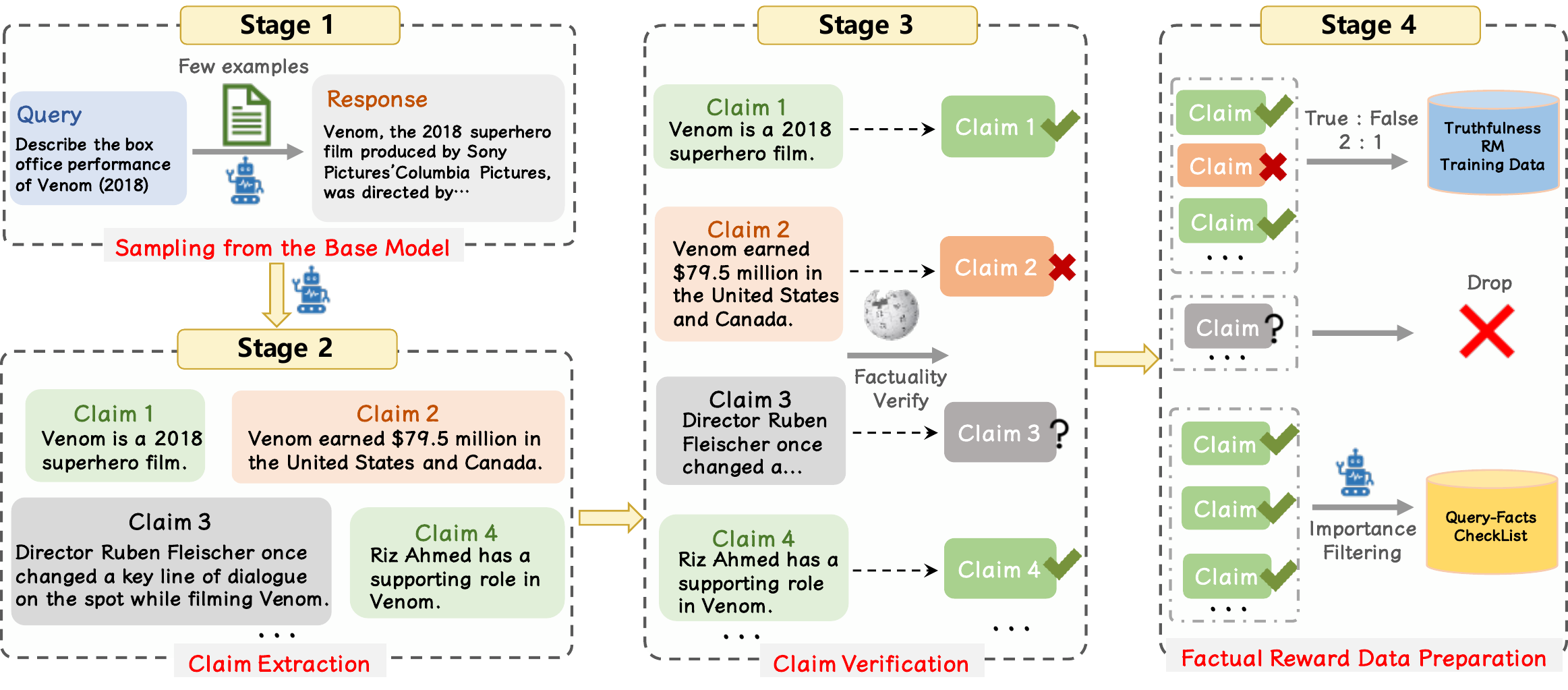}
		\caption{Offline data preparation pipeline. The process constructs the essential resources for knowledge-level consistency rewards—a factual checklist and truthfulness reward model training data—by extracting and verifying claims from the base model's responses.}
	\label{fig:data}
\end{figure}
The checklist $\Lambda(q)$ and the truthfulness reward model both rely on a practical approximation to $K_{\mathrm{vfe}}(q)$. We construct this approximation from three data sources, namely ELI5~\citep{fan2019eli5}, LongFact-Gen, and LongWiki-Gen, to ensure diversity and factual grounding. LongFact-Gen is regenerated using GPT-4.1 based on the original LongFact~\citep{wei2024long} prompts, and LongWiki-Gen is built by applying the methodology from~\citep{bang2025hallulens} to the GoodWiki corpus~\citep{GoodWiki}, ensuring no overlap with their respective test sets. The overall pipeline, illustrated in Fig.~\ref{fig:data}, consists of the following four stages.

\textbf{Stage 1: Sampling from the base model.} To probe what the base model can express for a query $q_i$, we prompt it to generate multiple responses in a few-shot setting. Specifically, for each query we perform $\nu$ independent samplings with a high temperature setting (in our experiments, $\nu=8$), yielding responses $\{o_{i,1},\ldots,o_{i,\nu}\}$ where $o_{i,j} \sim \pi_{\mathrm{base}}(\cdot \mid q_i,\mathcal{C})$ and $\mathcal{C}$ denotes the few-shot context. This step expands the observable expressed knowledge of the base model under the chosen prompting protocol.

\textbf{Stage 2: Claim extraction.} We then extract verifiable atomic claims from the sampled responses using a lightweight claim extraction model $f_{\mathrm{extract}}$. For each query $q_i$, we aggregate claims across all $\nu$ samples:
\begin{equation}
C(o_i) = \bigcup_{j=1}^{\nu} f_{\mathrm{extract}}(o_{i,j}) = \{c_{i,1},c_{i,2},\ldots,c_{i,M_i}\}.
\end{equation}
This converts unstructured generations into a structured candidate fact set that can be externally verified.

\textbf{Stage 3: External verification.} Each extracted claim $c$ is verified against a local Wiki20250716 knowledge index, built from a processed Wikipedia dump and indexed via~\citep{chen2017reading,wang2022text}. For each claim, the top-$k$ relevant documents $D(c)$ are retrieved (with $k=10$ in our experiments), and a verifier model $f_{\mathrm{verify}}$ assigns one of three labels:
\begin{equation}
l = f_{\mathrm{verify}}(c, D(c)), \qquad l \in \{\textsc{Support}, \textsc{Refute}, \textsc{Not Enough Info}\}.
\end{equation}
This step identifies which expressed claims are externally supported and therefore eligible to approximate the model's verified faithfully expressed knowledge.

\textbf{Stage 4: Reward resource construction.} Finally, the verified claims are curated into the two resources used during RL. First, the factual checklist $\Lambda(q_i)$ is built by aggregating all \textsc{Support}-labeled claims for query $q_i$ and then filtering duplicates and low-importance items, resulting in a concise blueprint of verified base-expressible facts. Second, the truthfulness reward model training set is built from \textsc{Support} and \textsc{Refute} claims. The former provides direct supervision for checklist-based recall and precision rewards, while the latter allows the truthfulness reward model to learn a retrieval-free approximation to the base model's reliable knowledge boundary.

\subsection{Reward Design and Policy Optimization}\label{sec:rewards}
We first formalize the key components involved in online RL. For a given query $q_i$, the policy model generates a structured response $o_i$ = \textless think\textgreater $\mathcal{T}_i$ \textless /think\textgreater~\textless answer\textgreater $\mathcal{A}_i$ \textless /answer\textgreater. Additionally, the factual checklist $\Lambda(q_i)$ is precomputed offline as described above. The final scalar reward is denoted by $R(o_i)$.

\subsubsection{Auxiliary Rewards}\label{sec:aux_reward}
To preserve overall generation quality beyond factuality, we introduce three lightweight auxiliary rewards. \textbf{General Reward} $R_g(\mathcal{A}_i)$ is provided by a preference reward model to maintain readability and human-preferred answer quality. \textbf{Format Reward} $R_f(o_i)$ enforces the required \texttt{<think>} and \texttt{<answer>} structure for reliable parsing. \textbf{Length Penalty} $R_l(\mathcal{A}_i)$ discourages excessively long or redundant generations and promotes concise, information-dense answers. Full details of these auxiliary rewards are provided in Appendix~\ref{sec:appendix_aux_reward}.

\subsubsection{Final Reward}\label{sec:reward_combination}
Our final reward combines the base-anchored KLC rewards with the auxiliary rewards:
\begin{equation}\label{eq:total}
R(o_i) =
\kappa \, R_{\text{recall}}(\mathcal{A}_i)
+ \gamma \, R_{\text{precision}}(\mathcal{A}_i)
+ \mu \, \widetilde{R}_{\text{truth}}(\mathcal{A}_i)
+ 0.1 \, R_g(\mathcal{A}_i)
+ R_l(\mathcal{A}_i)
+ R_f(o_i),
\end{equation}
where $\widetilde{R}_{\text{truth}}$ denotes either $R_{\text{truth}}$ or $R_{\text{truth}}^{\text{variant}}$, and $\kappa,\gamma,\mu \in [0,1]$ satisfy $\kappa + \gamma + \mu = 1$. In this formulation, $R_{\text{recall}}$ promotes the recall of verified base knowledge, $R_{\text{precision}}$ improves checklist-scope accuracy, and $\widetilde{R}_{\text{truth}}$ suppresses unfaithful claims beyond the checklist. Checklist-only, truth-only, and full factual reward variants are all obtained by setting the corresponding coefficients. We discuss the choice of $\kappa$, $\gamma$, and $\mu$ in Section~\ref{sec:ablation}.

\subsubsection{Policy Optimization}\label{sec:po}
To optimize the policy toward the reward objective above, we employ Group Relative Policy Optimization (GRPO)~\citep{shao2024deepseekmath}. For a given query $q$, the behavior policy $\pi_{\theta_{\mathrm{old}}}$ generates a group of $G$ responses $\{o_i\}_{i=1}^{G}$. The advantage of the $i$-th response is computed by normalizing the group-level rewards $\{R(o_i)\}_{i=1}^{G}$:
\begin{equation}
\hat{A}_{i,t} =
\frac{R(o_i) - \mathrm{mean}(\{R(o_i)\}_{i=1}^{G})}
{\mathrm{std}(\{R(o_i)\}_{i=1}^{G})},
\qquad
r_{i,t}(\theta) =
\frac{\pi_\theta(o_{i,t} \mid q, o_{i,<t})}
{\pi_{\theta_{\mathrm{old}}}(o_{i,t} \mid q, o_{i,<t})}.
\end{equation}
When the within-group standard deviation is zero, we set $\hat{A}_{i,t}=0$. The GRPO objective is then
\begin{equation}
\mathcal{J}_{\mathrm{GRPO}}(\theta) =
\frac{1}{G} \sum_{i=1}^{G} \frac{1}{|o_i|} \sum_{t=1}^{|o_i|}
\left(
\min\!\Big(
r_{i,t}(\theta)\hat{A}_{i,t},
\mathrm{clip}(r_{i,t}(\theta), 1-\epsilon, 1+\epsilon)\hat{A}_{i,t}
\Big)
- \beta D_{\mathrm{KL}}(\pi_\theta \| \pi_{\mathrm{ref}})
\right)
\end{equation}
where $\epsilon$ is the clipping range, $\beta$ controls the KL penalty, and $\pi_{\mathrm{ref}}$ is the reference model.

%% file: experiment.tex
\section{Experiments}
\subsection{Experimental Settings}
\textbf{Dataset and Metrics.} We evaluate model performance on four long-form factual benchmarks: FActScore~\citep{min2023factscore}, Hallulens-LongWiki (abbreviated as LongWiki)~\citep{bang2025hallulens}, LongFact~\citep{wei2024long}, and Factory~\citep{chen2025factory}. Metrics include FActScore, Recall@K, Precision, F1@K and WR (Win Rate). Detailed descriptions of the training data processing pipeline are provided in Section~\ref{sec:rl_data} and the Appendix~\ref{sec:appendix_data}. For a detailed description of the evaluation metrics and related configuration settings, please refer to Appendix~\ref{sec:appendix_evaluation}.

\textbf{Models and Baselines.} To comprehensively evaluate the effectiveness of our proposed method and ensure a fair and thorough comparison across different training paradigms and model scales, we conduct extensive experiments on Qwen2.5~\citep{Yang2024Qwen25TR} family models of various sizes, including 7B, 14B, and 32B. Our baselines include a wide range of representative methods: pretrained base models (Base), prompting method CoVe~\citep{dhuliawala2023chain}, supervised fine-tuned models (DeepSeek Distillation Series~\citep{guo2025deepseek}), unsupervised method Intuitior~\citep{zhao2025learning}, Self-Eval-P(True)~\citep{zhang2024self} (leveraging P(True) to refine factual precision), DPO~\citep{rafailov2023direct} and GRPO~\citep{shao2024deepseekmath} variants based on different reward signals (DPO+FActScore\&KLC rewards, GRPO+FActScore\&KLC rewards, and GRPO+FActScore+No. claims).

\textbf{Training Details.} Training is conducted on 16$\times$H100 (80GB) GPUs with full-parameter tuning. For all experiments, we employ a consistent learning rate of 1e-6 with cosine warmup. Further implementation details are provided in the Appendix~\ref{sec:appendix_train_parameters}.

\subsection{Main Results}
As summarized in Table~\ref{tab:main}, our method achieves significant improvements under both training paradigms. KLCF-zero, trained directly from the base model without SFT, attains a higher performance ceiling than SFT-based approaches—we attribute this to avoiding the knowledge forgetting and alignment tax introduced by SFT-based distillation~\citep{luo2025empirical,fu-etal-2024-disperse}. Compared to FActScore-based reward methods (including GRPO+FActScore trained from the base model), which consistently sacrifice recall for precision, KLCF achieves the best F1 balance across benchmarks via dual-fact alignment. For detailed analysis of reward trends and training dynamics, please refer to Appendix~\ref{sec:appendix_main_exp}.

\begin{table}[t]
\centering
\setlength{\tabcolsep}{1.5pt}
\renewcommand{\arraystretch}{1.2} %
\scalebox{0.8}{
\begin{tabular}{c|c|ccc|cccc|cccc}
\toprule
\multirow{2}{*}{\textbf{Model}} & \multicolumn{1}{c|}{\textbf{FActScore}} & \multicolumn{3}{c|}{\textbf{LongWiki}} & \multicolumn{4}{c|}{\textbf{LongFact}} & \multicolumn{4}{c}{\textbf{Factory}} \\
\cline{2-13}
& FS & R@32 & Prec & F1@32 & R@64 & Prec & F1@64 & WR & R@64 & Prec & F1@64 & WR\\
\hline
\multicolumn{13}{c}{\textbf{Training from Base Model}} \\
\hline
Base (10-shot)     & 46.8\% & 0.661 & 0.435 & 0.511 & 0.614 & 0.685 & 0.637 & - & 0.239 & 0.302 & 0.262 & - \\
+ CoVe   & 46.0\%   & 0.435   & 0.381   & 0.368 & 0.380 & 0.676 & 0.452 &19.8\% & 0.179
& 0.304 & 0.206 & 17.9\% \\
+ GRPO\&FActScore   & 54.8\%   & 0.532   & 0.473   & 0.538 & 0.634 & 0.690 & 0.663 &54.7\% & 0.253
& 0.318 & 0.278 & 35.9\% \\
+ Self-Eval-P(True) & 52.0\%	& 0.498	& \textbf{0.567}	& 0.504	& 0.232	& \textbf{0.718}	& 0.338 & 8.63\% & 0.126 & 0.337 & 0.173 & 25.9\% \\
\rowcolor{lightgray}
+ KLCF-zero (Ours)   & \textbf{61.2\%}  & \textbf{0.681} & 0.552 & \textbf{0.568} & \textbf{0.776} & 0.704 & \textbf{0.733} & \textbf{94.6\%} &  \textbf{0.296} & \textbf{0.350} & \textbf{0.309} & \textbf{65.5\%} \\

\hline
\multicolumn{13}{c}{\textbf{Training from SFT Model}} \\
\hline

SFT (Distill)    & 48.7\%   & 0.503   & 0.532   & 0.494 & 0.403 & 0.760 & 0.507 & - &  0.138 & 0.286 & 0.175 & - \\

+ Intuitor   & 52.7\% & 0.505 & 0.580 & 0.512 & 0.397 & 0.774 & 0.502 & 41.5\% & 0.145 & 0.294 & 0.183 & 41.8\% \\
+ DPO\&FActScore   & 53.0\%  & 0.502  & 0.538  & 0.496 & 0.407 & 0.765 & 0.512 & 47.2\% & 0.158 & 0.308 & 0.190 & 44.4\% \\
+ DPO\&KLC   & 54.9\%  & 0.505  & 0.584  & 0.517 & 0.432 & 0.779 & 0.530 & 56.9\% & 0.157 & 0.312 & 0.186 & 52.4\% \\
+ GRPO\&FActScore & \textbf{57.6\%}  & 0.472  & \textbf{0.647}  & 0.510 & 0.406 & 0.780 & 0.504 & 29.6\% & 0.143 & 0.351 & 0.184 & 30.2\% \\
+ GRPO\&FActScore\&No.Claims & 53.9\%	& 0.616	& 0.568	& 0.610	& 0.652 & 0.757	& 0.685	& 70.3\% & 0.213 & 0.361 & \textbf{0.247} & 61.5\% \\
\rowcolor{lightgray}
+ KLCF (Ours)   & 55.7\%  & \textbf{0.663} & 0.597 & \textbf{0.651} & \textbf{0.668} & \textbf{0.783} & \textbf{0.692} & \textbf{79.3\%} & \textbf{0.221} & \textbf{0.383} & 0.242 & \textbf{66.5\%} \\

\bottomrule
\end{tabular}
}
\captionsetup{width=0.9\linewidth}
\caption{Results on Qwen2.5-14B. SFT denotes the DeepSeek-R1-Distill-Qwen-14B. R@32 and Prec are Recall@32 and Precision. Best scores are bolded.}
\label{tab:main}
\end{table}

\subsection{Ablation Study}\label{sec:ablation}
As shown in Table~\ref{tab:ablation}, ablating the Dual-Fact Alignment mechanism validates its critical role. Using only the Checklist Reward  ($\kappa=1/3,\gamma=2/3,\mu=0$) yields high recall but low precision, as excessive coverage introduces errors. Conversely, only the Truthfulness Reward ($\kappa=0,\gamma=0,\mu=1$) boosts precision at a severe cost to recall, indicating over-conservatism. Systematically adjusting the reward weights (from top to bottom in Table~\ref{tab:ablation}) reveals a clear trade-off: reducing the Checklist weight while increasing the Truthfulness weight monotonically improves precision but consistently degrades recall. This trend highlights the limitation of optimizing either objective in isolation. By integrating both rewards, KLCF achieves a superior balance. The Truthfulness Reward suppresses hallucinations encouraged by the Checklist Reward, which in turn counteracts conservatism, demonstrating the necessity of our dual mechanism.

\begin{table}[t]
\centering
\setlength{\tabcolsep}{1.5pt}
\renewcommand{\arraystretch}{1.2} %
%\CheckmarkBold & \XSolidBrush
\scalebox{0.85}{
\begin{tabular}{ccc|c|ccc|cccc|cccc}
\toprule
\multicolumn{3}{c|}{\textbf{Reward Weight}} & \multicolumn{1}{c|}{\textbf{FActScore}} & \multicolumn{3}{c|}{\textbf{LongWiki}} & \multicolumn{4}{c|}{\textbf{LongFact}} & \multicolumn{4}{c}{\textbf{Factory}} \\
\cline{1-15}

\makecell[c]{Recall\\$\kappa$} & \makecell[c]{Precision\\$\gamma$} & \makecell[c]{Truth\\$\mu$} & FS & R@32 & Prec & F1@32 & R@64 & Prec & F1@64 & WR & R@64 & Prec & F1@64 & WR \\
\hline

0 & 0 & 0 & 46.8\% & 0.661 & 0.435 & 0.511 & 0.614 & 0.685 & 0.637 & - & 0.239 & 0.314 & 0.262 & - \\

$\frac{1}{3}$ & $\frac{2}{3}$ & 0 & 48.8\% &  0.678 & 0.424 & 0.508 & 0.729 & 0.672 & 0.695 & 95.0\% & 0.243 & 0.266 & 0.248 & \textbf{81.5\%} \\

$\frac{1}{3}$ & $\frac{1}{3}$ & $\frac{1}{3}$ & 55.7\%  & \textbf{0.695} & 0.490 & 0.545 & 0.748 & 0.689 & 0.713 & \textbf{96.4\%} & 0.273 & 0.258 & 0.290 & 79.3\%\\

\rowcolor{lightgray}
$\frac{1}{4}$ & $\frac{1}{4}$ & $\frac{1}{2}$ & 61.2\%  & 0.681 & 0.552 & \textbf{0.568} & 0.776 & 0.704 & 0.733 & 94.6\% & \textbf{0.296} & 0.350 & \textbf{0.309} & 65.5\% \\

0.4 & 0.2 & 0.4 & 57.5\% & 0.692 & 0.512 & 0.553 & \textbf{0.781} & 0.706 & \textbf{0.737} & 95.4\% & 0.281 & 0.323 & 0.292 & 70.3\% \\

0 & $\frac{1}{3}$ & $\frac{2}{3}$ & 69.5\% & 0.434 & 0.679 & 0.464 & 0.642 & 0.761 & 0.685 & 49.2\% & 0.241 & 0.435 & 0.290 & 22.9\% \\

0 & 0 & 1 & \textbf{69.6\%} & 0.369 & \textbf{0.724} & 0.422 & 0.633 & \textbf{0.793} & 0.685 & 36.8\% & 0.243 & \textbf{0.492} & 0.296 & 15.3\% \\
\bottomrule
\end{tabular}}
\captionsetup{width=0.9\linewidth}
\caption{Ablation study on reward components. Results are for Qwen2.5-14B under the RL-zero (training from base model) setting. The first row is the base model (10-shot) reference, and the gray row denotes our optimal reward weighting.}
\label{tab:ablation}
\end{table}

\subsection{Scaling Study}
To verify the generalizability of the KLCF method across different model scales, we conduct scaling experiments on Qwen2.5-7B and 32B models. As shown in Table~\ref{tab:scaling}, for both the 7B and 32B models, our proposed method under the ``Training from Base Model'' setting significantly outperforms the original base model and the CoVe prompting approach across all factuality metrics. The results demonstrate that the performance improvement brought by KLCF is consistent regardless of model scale. Our method effectively enhances long-form factuality in small (7B), medium (14B), and large (32B) models, confirming that the framework possesses strong scalability and generalizability.

\begin{table}[t]
\centering
\setlength{\tabcolsep}{1.5pt}
\renewcommand{\arraystretch}{1.2} %

\scalebox{0.85}{
\begin{tabular}{c|c|ccc|cccc|cccc}
\toprule
\multirow{2}{*}{\textbf{Model}} & \multicolumn{1}{c|}{\textbf{FActScore}} & \multicolumn{3}{c|}{\textbf{LongWiki}} & \multicolumn{4}{c|}{\textbf{LongFact}} & \multicolumn{4}{c}{\textbf{Factory}} \\
\cline{2-13}
& FS & R@32 & Prec & F1@32 & R@64 & Prec & F1@64 & WR & R@64 & Prec & F1@64 & WR \\
\hline
\multicolumn{13}{c}{\textbf{Qwen2.5-7B}} \\
\hline
Base (10-shot)  & 36.6\% & 0.529 & 0.359 & 0.412 & 0.522 & 0.600 & 0.548 & - & 0.203 & 0.267 & 0.223 & - \\
+ CoVe   & 40.1\% & 0.293 & 0.357 & 0.294 & 0.326 & 0.605 & 0.383 & 15.5\% & 0.171 & 0.278 & 0.192 & 11.8\% \\
\rowcolor{lightgray}
+ KLCF-zero (Ours)   & \textbf{55.0\%}  & \textbf{0.633} & \textbf{0.487} & \textbf{0.514} & \textbf{0.724} & \textbf{0.675} & \textbf{0.693} & \textbf{94.8\%} & \textbf{0.258} & \textbf{0.301} & \textbf{0.271} & \textbf{78.9\%} \\

\hline
\multicolumn{13}{c}{\textbf{Qwen2.5-32B}} \\
\hline

Base (10-shot)     & 40.0\% & 0.664 & 0.453 & 0.520 & 0.531 & 0.691 & 0.575 & - & 0.234 & 0.318 & 0.256 & - \\
+ CoVe   & 45.9\% & 0.375 & 0.410 & 0.355 & 0.379 & 0.718 & 0.466 & 19.5\% & 0.191 & 0.346 & 0.229 & 15.3\% \\
\rowcolor{lightgray}
+ KLCF-zero (Ours)   & \textbf{59.3\%} & \textbf{0.812} & \textbf{0.532} & \textbf{0.631} &  \textbf{0.784} & \textbf{0.747} & \textbf{0.760} & \textbf{96.2\%} & \textbf{0.307} & \textbf{0.434} & \textbf{0.322} & \textbf{71.3\%} \\

\bottomrule
\end{tabular}
}
\captionsetup{width=0.9\linewidth}
\caption{Scaling study of KLCF on Qwen2.5-7B and 32B models under the ``Training from Base Model'' setting.}
\label{tab:scaling}
\end{table}

\subsection{Generalization to Non-Thinking Models}
While our primary experiments demonstrate the effectiveness of KLCF starting from a base model under a thinking-style architecture, we further investigate its generalization capability by applying it to a standard non-thinking model. For this experiment, we adapt the reward function by removing the format reward (as it is unnecessary for a standard conversational model) and the general reward (relying instead on the KL divergence penalty to prevent excessive deviation), while retaining the length penalty to control verbosity. Results show that KLCF still achieves a substantial improvement in factuality metrics (as shown in Table~\ref{tab:non_thinking}), confirming that its effectiveness is not reliant on a thinking-style architecture and generalizes robustly to conventional conversational models.

\begin{table}[h]
\centering
\setlength{\tabcolsep}{1.5pt}
\renewcommand{\arraystretch}{1.2} %
%\CheckmarkBold & \XSolidBrush
\scalebox{0.85}{
\begin{tabular}{c|c|ccc|cccc|cccc}
\toprule
 & \multicolumn{1}{c|}{\textbf{FActScore}} & \multicolumn{3}{c|}{\textbf{LongWiki}} & \multicolumn{4}{c|}{\textbf{LongFact}} & \multicolumn{4}{c}{\textbf{Factory}} \\
\cline{2-13}

 & FS & R@32 & Prec & F1@32 & R@64 & Prec & F1@64 & WR & R@64 & Prec & F1@64 & WR \\
\hline

Qwen2.5-14B-Instruct & 61.1\% & 0.548 & 0.548 & 0.578 & 0.412 & 
0.788 & 0.531 & - & 0.193 & 0.388 & 0.245 & - \\
\rowcolor{lightgray}

+ KLCF (Ours) & \textbf{65.0\%} & \textbf{0.708} & \textbf{0.580} & \textbf{0.595} & \textbf{0.686} & \textbf{0.808} & \textbf{0.747} & \textbf{97.6\%} & \textbf{0.290} & \textbf{0.398} & \textbf{0.306} & \textbf{67.1\%} \\

\bottomrule
\end{tabular}}
\captionsetup{width=0.9\linewidth}
\caption{Performance improvement on factuality benchmarks after adapting KLCF to a non-thinking, instruction-tuned model (Qwen2.5-14B-Instruct).}
\label{tab:non_thinking}
\end{table}

\subsection{Truthfulness Reward Analysis}
We conducted a comparative analysis between the standard truthfulness reward calculation (Eq.~\ref{eq:truth}) and its variant (Eq.~\ref{eq:truth_var}) that leverages the pre-verified factual checklist to reduce evaluation noise. Both methods enhance factuality, but they exhibit distinct characteristics in balancing precision and recall. Due to space constraints, a detailed discussion of the results and their implications is provided in Appendix~\ref{sec:appendix_truth}.

\subsection{Efficiency Analysis}\label{exp:efficiency}
\begin{table}[h]
\centering
\setlength{\tabcolsep}{1.5pt}
\renewcommand{\arraystretch}{1.2} %

\scalebox{0.90}{
\begin{tabular}{c|c|c >{\columncolor[gray]{0.9}}c >{\columncolor[gray]{0.9}}c ccc}
\toprule
& Reward & Time (s) &  $\uparrow$(vs. FActScore)  & $\uparrow$(vs. VeriScore) & \# Searches & Input Tokens & Output Tokens \\
\hline
% \cline{2-8}
\multirow{4}{*}{Serial} 
& FActScore & 198.06 & 1.00x & 1.18x & 51.74 & 55666.20 & 5701.56 \\
& VeriScore & 234.50 & 0.84x & 1.00x & 32.64 & 335909.78 & 3442.96 \\
& Ours & 57.29 & \textbf{3.46x} & \textbf{4.09x} & 0 & 1457.76 & 855.24 \\
\hline
\multirow{4}{*}{Parallel} 
& FActScore & 10.58 & 1.00x & 1.03x & 52.02 & 56677.50 & 5926.04 \\
& VeriScore & 10.90 & 0.97x & 1.00x & 32.56 & 334077.92 & 3407.28 \\
& Ours & 2.02 & \textbf{5.24x} & \textbf{5.40x}  & 0 & 1452.62 & 851.46 \\
% \rowcolor{lightgray}
\bottomrule
\end{tabular}
}
\captionsetup{width=0.9\linewidth}
\caption{Efficiency comparison of different factual reward calculation methods. All reported values are averaged over 50 samples. Our proposed reward (Checklist (recall + precision) + Truthfulness) achieves significant speedups in both serial and parallel modes. The ``\# Searches'' column indicates the number of times the method called the external search engine API. Notably, our method requires no such calls and maintains low input \& output token consumption.}
\label{tab:efficiency}
\end{table}
To evaluate the efficiency of our reward design, we compare the computational cost of various factuality reward methods. Experimental details are provided in Appendix~\ref{sec:appendix_efficiency}. As summarized in Table~\ref{tab:efficiency} and Table~\ref{tab:pipeline-comparison}, our KLC rewards achieves a \textbf{$3.46\times$} and \textbf{$4.09\times$} speedup over FActScore and VeriScore in serial mode, which further increases to \textbf{$5.24\times$} and \textbf{$5.40\times$} under parallel execution. More importantly, the detailed breakdown in Table~\ref{tab:pipeline-comparison} shows that our online reward computation takes only \textbf{127 seconds} per training step, compared to \textbf{1056 seconds} for FActScore—a \textbf{5.4×} reduction. While we incur a one‑time offline preparation cost (20 hours) to construct checklists and train reward models, this investment is \textbf{permanently reusable} across multiple RL runs and model scales. In contrast, FActScore‑based methods rely on external search APIs, which suffer from rate limits, network instability, and high latency.

\subsection{Case Study}\label{exp:case
-study}
In this section, we present a case study to qualitatively illustrate the impact of our KLCF framework. We compare the responses of the Qwen2.5-14B-Instruct model and its KLCF-enhanced version to a specific query from the LongFact test set. The results are shown in Table~\ref{tab:case_study}. Our analysis shows that the KLCF-trained model produces a significantly longer and more comprehensive response, indicating a substantial improvement in factual recall by utilizing more of its internal knowledge. Furthermore, the enhanced output also demonstrates higher factual precision, with fewer observable errors compared to the base model. This dual improvement highlights the effectiveness of our approach in mitigating hallucinations while reducing over-conservatism. For a detailed analysis, including the specific query and the model outputs with annotated errors, please refer to Appendix~\ref{sec:appendix_case_study}.

%% file: related_work.tex
\section{Related Work}
\subsection{Long-Form Factuality Evaluation}
Fine-grained factuality assessment of generated text serves as a fundamental prerequisite for reliable factuality alignment. Early approaches typically decompose long-form text into atomic facts and verify them against external knowledge to evaluate factual accuracy. FActScore~\citep{min2023factscore} and FacTool~\citep{chern2023factool} establish this paradigm, employing retrieval and large verification models to determine the veracity of each atomic fact. Subsequently, methods such as SAFE~\citep{wei2024long} and VeriScore~\citep{song2024veriscore} further refine the fact extraction and verification pipeline, while numerous follow-up studies~\citep{ni2024afacta,bishop2024longdocfactscore,wanner2025dndscore} continue to advance this direction by improving evaluation effectiveness and efficiency. However, these methods generally rely on external retrieval interfaces and large-scale verification models, resulting in high computational overhead and making them unsuitable for training scenarios in online reinforcement learning that require high-frequency reward signals.

\subsection{Long-Form Factuality Alignment}
To directly enhance the factuality of LLMs, researchers have proposed various alignment methods. Among approaches based on supervised fine-tuning (SFT) and direct preference optimization (DPO), methods such as FactTune-FS~\citep{tian2023fine} and FLAME~\citep{lin2024flame} leverage scores from evaluators like FActScore to guide models toward factual preferences, while FactAlign~\citep{huang2024factalign} and Mask-DPO~\citep{gu2025mask} construct preference pairs based on external retrieval for DPO training. However, these methods are generally confined to offline training scenarios and often prioritize factual precision at the expense of recall, leading to overly conservative model behavior.

In recent years, a growing body of work has begun incorporating factuality signals as rewards within RL frameworks. For example, KnowRL~\citep{ren2025knowrl} introduces factuality rewards based on knowledge verification during training to help the model recognize its knowledge boundaries, and TruthRL~\citep{wei2025truthrl} proposes a triple reward mechanism that distinguishes between correct answers, hallucinations, and abstentions, encouraging the model to refrain from responding when uncertain to improve truthfulness. Additionally, several other studies~\citep{chen2025learning,zhang2025reinforcement,li2025hallucination} explore the use of various types of factuality signals as rewards to enhance model capability and reduce hallucinations. Nevertheless, most of these methods still rely on time-consuming external retrieval processes, which limits their applicability in large-scale online RL training. In contrast to the above works, the proposed KLCF framework designs a fully external-knowledge-free and lightweight reward mechanism. Through dual-fact alignment, KLCF jointly optimizes factual recall and precision, effectively enhancing factuality while avoiding the decline in expressiveness caused by excessive conservatism.

\subsection{RL Without External Feedback}
To overcome the limitations of verifiers, researchers are increasingly turning to reward signal generation methods that operate without external feedback. These label-free RL approaches primarily develop along two representative directions. One line of research focuses on deriving reward signals from models' internal confidence estimates, enhancing predictive certainty by rewarding low-entropy or self-consistent outputs~\citep{zhao2025learning,zhang2025consistent,shafayat2025can,agarwal2025unreasonable,li2025jointly,prabhudesai2025maximizing}. For instance, Intuitor~\citep{zhao2025learning} leverages the model's own confidence estimations as unsupervised optimization signals to effectively improve the factuality of generated content. The other predominant paradigm constructs supervisory signals through collective decision-making mechanisms~\citep{zhang2025right,zuo2025ttrl}, as exemplified by TTRL~\citep{zuo2025ttrl}, which aggregates majority-voted answers from multiple generated samples to serve as pseudo-ground-truth labels for reinforcement learning updates.

\subsection{Alignment Tax}
The alignment tax is a prevalent phenomenon in RLHF and SFT processes. Some studies~\citep{ouyang2022training,huang2025safety,lin2024mitigating} indicate that alignment often leads models to become ``overly conservative'', sacrificing capability for improved safety. Some models like the DeepSeek-R1-Distill series perform well on general instruction-following tasks, their performance on long-form factuality benchmarks often falls short of the original base models, serving as a concrete manifestation of the alignment tax (see Table~\ref{tab:main}). More critically, current FActScore-guided RL methods further exacerbate this issue: to avoid negative judgments from external verifiers, models adopt a ``less-is-better" generation strategy, leading to a substantial decline in recall and ultimately reducing the overall F1 score. Therefore, a key motivation of this work is to mitigate the alignment tax by conducting RL directly from the base model with knowledge-level consistency as the objective. This approach enables the model to fully express its parametric knowledge while avoiding content that exceeds its knowledge boundaries.

%% file: conclusion.tex
\section{Conclusion, Limitations, and Future Work}\label{sec:conclusion}
We propose KLCF, a reinforcement learning framework that reduces hallucination in long-form generation by aligning the policy’s expressed knowledge distribution with the base model’s parametric knowledge boundary. The problem is formalized as a constrained distribution matching objective: maximizing recall of high‑probability facts while strictly forbidding generation outside the support of the base knowledge. Our Dual‑Fact Alignment mechanism realizes this objective through jointly optimized precision and recall terms. Experiments show that KLCF significantly surpasses existing baselines, and its retrieval‑free, lightweight reward design enables efficient scaling to large‑scale online RL training.

\textbf{Limitations and Future Work.} KLCF currently operates only in a closed‑book QA setting, with factuality rewards applied only at the response level, lacking fine‑grained supervision over intermediate reasoning. Future work will explore step‑wise factual alignment, such as process‑based rewards at the sentence level or within the chain of thought. Furthermore, while fact checklist construction currently relies on a static knowledge source, the framework can be naturally extended to incorporate real‑time retrieval or human‑annotated data. 

Additionally, KLCF constrains generation to the base model's knowledge boundary, raising two limitations. First, the base model's parametric knowledge may be incomplete or erroneous; using its boundary as a factuality proxy could reinforce these mistakes. To mitigate this, our truthfulness reward model is trained with external Wikipedia knowledge to correct potential errors. Second, this constraint prevents introducing new knowledge beyond the model's parameters—a deliberate design choice to make the model honestly express what it has already learned, not to learn new facts. If external knowledge is needed, the framework can be extended with retrieval or human-annotated data. We view these not as unsolvable problems, but as an explicit trade‑off between honesty and correctness, which future work can further alleviate by incorporating external knowledge sources.

%% file: appendix.tex
\section{Implementation Details}\label{sec:appendix_details}
\subsection{Data Construction}\label{sec:appendix_data}
\textbf{RL Training Dataset.} ELI5~\citep{fan2019eli5} is a publicly available long-form question-answering dataset, from which we select 7993 samples for training. To avoid overlap with evaluation benchmarks, we construct two new datasets adhering to established methodologies: LongFact-Gen, containing 4348 samples generated by reproducing the original LongFact~\citep{wei2024long} prompt strategy using GPT-4.1; and LongWiki-Gen, comprising 2248 samples built from the GoodWiki~\citep{GoodWiki} corpus following the procedure described in HalluLens~\citep{bang2025hallulens}. Both training sets are distinct from their corresponding test benchmarks.

We then follow the data preprocessing pipeline outlined in Section~\ref{sec:rl_data} to perform uniform cleaning and formatting on all three datasets. The training datasets are summarized in the table \ref{tab:train_data_analysis}.

\begin{table}[h]
    \centering
    \renewcommand{\arraystretch}{1.2} %
    \scalebox{0.9}{
    \begin{tabular}{c|ccc|c}  %c|ccc|
    \toprule
    & \multicolumn{3}{c|}{\textbf{RL}} & \multicolumn{1}{c}{\textbf{DPO}} \\
\cline{2-5}
     & \textbf{Dataset Size} & \textbf{Avg. True Claims} & \textbf{Avg. Checklists} & \textbf{Dataset Size} \\
    \hline
    7B  & 12680 & 34.76  & 10.09 & 12680 \\
    14B & 13230  & 38.34  & 11.03 & 13230 \\
    32B & 12967  &  37.19  & 11.10 & 12967 \\
    \bottomrule
    \end{tabular}}
    \caption{Statistics of training data.}
    \label{tab:train_data_analysis}
\end{table}

\textbf{DPO Training Dataset.} In this work, we employ the DeepSeek-R1-Distill-Qwen model for DPO training. To construct the DPO training dataset, we first sample 8 responses for each query from the RL dataset using three different sizes of models (7B, 14B and 32B). Subsequently, we extract claims from all responses for every query using a lightweight trained claim extraction model along with Prompt~\ref{prompt:claim_extract}. We then verify all claims of each response against a locally built Wiki20250716 index, utilizing the Qwen2.5-72B-Instruct and Prompt~\ref{prompt:claim_verify}, to obtain the FActScore for every response. Finally, we form the \textless chosen, rejected\textgreater~ pairs by selecting the highest and lowest-scoring responses for each query, thereby completing the preparation of the DPO dataset.

Similarly, to construct DPO training data based on the reward mechanism proposed in this paper, we need to assign a reward value to each response generated for every query, according to the reward computation method defined in Section~\ref{sec:klc_obj}. Specifically, we first use a lightweight checklist verification model with Prompt~\ref{prompt:checklist_verify} to extract the checklist verification results for each response, and then compute the reward $R_{\text{recall}}$ and the reward $R_{\text{precision}}$ based on Eq.~(\ref{eq:recall}) and Eq.~(\ref{eq:precision}), respectively. Subsequently, based on the previously extracted claims, we employ a truthfulness reward model with Prompt~\ref{prompt:p_true} to obtain the credibility probability score for each claim, and calculate the reward $R_{\text{truth}}$ according to Eq.~(\ref{eq:truth}). Finally, for each query, we select the responses with the highest and lowest reward values to form the \textless chosen, rejected\textgreater~ preference pairs. The training datasets used for DPO are summarized in the table \ref{tab:train_data_analysis}.

\textbf{Test Dataset.} In this paper, we conduct a comprehensive evaluation of the trained models on four public long-form text evaluation benchmarks: FActScore~\citep{min2023factscore}, Hallulens-LongWiki~\citep{bang2025hallulens}, LongFact~\citep{wei2024long}, and Factory~\citep{chen2025factory}. The statistics for these datasets are detailed in Table~\ref{tab:test_data_analysis}. Specifically, for FActScore, we use 500 samples provided in its official release as the test set. For Hallulens-LongWiki, we strictly follow the methodology described in the original paper to generate 250 test samples. For the LongFact benchmark, we select 250 samples from its ``Objects'' subset. For the Factory benchmark, we randomly select 250 samples from its ``Hard'' subset to form the test set.

\begin{table}[h]
    \centering
    \setlength{\tabcolsep}{1.6pt}
    \scalebox{0.9}{
    \begin{tabular}{c|cccc}  %c|ccc|
    \toprule
    & \textbf{FActScore} & \textbf{Hallulens-LongWiki} & \textbf{LongFact-Objects} & \textbf{Factory-Hard}\\
    \hline
    Total Dataset Size  & 500 & -  & 1140 & 421 \\
    \hline
    Test Dataset Size  & 500 & 250  & 250 & 250 \\
    \bottomrule
    \end{tabular}}
    \caption{Statistics of test dataset.}
    \label{tab:test_data_analysis}
\end{table}

\subsection{Baselines}
This section provides a detailed description of the baseline methods included in our experiments to ensure a comprehensive and fair comparison. Our evaluation encompasses a diverse set of representative approaches across different training paradigms.

The \textbf{Base} model refers to the pretrained Qwen2.5 models (7B, 14B, 32B) evaluated in a 10-shot setting, serving as the foundational performance benchmark. The prompting-based baseline is represented by \textbf{CoVe}~\citep{dhuliawala2023chain}, which reduces hallucinations through a self-verification mechanism during inference without updating model parameters. For supervised fine-tuning, we include the \textbf{DeepSeek Distillation Series}~\citep{guo2025deepseek}. We also compare against Intuitor~\citep{zhao2025learning}, an unsupervised method that improves factuality by leveraging the model's intrinsic confidence estimates.

Furthermore, we incorporate several reinforcement learning baselines to isolate the contribution of our reward design. This includes DPO and GRPO optimized with different reward signals: \textbf{DPO + FActScore} and \textbf{GRPO + FActScore} use the external FActScore metric as the reward signal, favoring precision but often at the cost of recall. In contrast, \textbf{DPO + KLC Rewards} and \textbf{GRPO + KLC Rewards} (which is our full KLCF method) are trained using the knowledge-level consistency rewards introduced in this work, enabling joint optimization of recall and precision through dual-fact alignment.

\subsection{Training Setups}\label{sec:appendix_train_parameters}
In this section, we systematically elaborate on the hyperparameter configurations employed across all core experiments in this paper, covering training stages such as Supervised Fine-Tuning (SFT), Direct Preference Optimization (DPO), and Reinforcement Learning (RL). To ensure the reproducibility and consistency of our experiments, hyperparameter settings are kept uniform within the same training stage. For the comparative model Intuitor, we adopt the relevant configurations from its original paper. All specific parameter values are summarized in Table~\ref{tab:training_parameters}.

\begin{table}[t]
    \centering
    \fontsize{10}{15}\selectfont %
    \setlength{\tabcolsep}{1.5pt}
    \renewcommand{\arraystretch}{4} %
    \scalebox{0.65}{
    \begin{tabular}{>{\centering}p{2.3cm}|>{\centering}p{2cm}|c|c|c|c}
    \toprule

    \multicolumn{6}{c}{\textbf{RL and DPO}} \\
    \hline
    
    \textbf{task} & \textbf{framework} & \textbf{data} & \textbf{actor\_rollout\_ref} & \textbf{algorithm} & \textbf{trainer} \\
\cline{1-6}
    RL & VeRL
    & \makecell[c]{
        max\_prompt\_length=8192 \\
        max\_response\_length=4096 \\
        train\_batch\_size=64
    }
    & \makecell[c]{
        entropy\_coeff=0.0 \\
        use\_kl\_loss=False \\
        kl\_loss\_coef=0.0 \\
        rollout\_n=8 \\
        ppo\_epochs=1 \\
        ppo\_mini\_batch\_size=64 \\
        lr=1e-6,weight\_decay=0.1 \\
        lr\_warmup\_steps\_ratio=0.1 \\
        warmup\_style=cosine \\
        temperature=1.0, top\_p=0.05 \\
        gpu\_memory\_utilization=0.5 \\
        tensor\_model\_parallel\_size=4 \\
        gradient\_checkpointing=True \\
    }
    & \makecell[c]{
        gamma=1.0 \\
        lam=1.0 \\
        $L_m$=2048 \\
        $L_m-L_c$=850 \\
    }
    & \makecell[c]{
        total\_epochs=2 \\
        nnodes=2 \\
        n\_gpus\_per\_node=8 \\
        save\_freq=20
    }
    \\  %
    \hline
    Intuitor & VeRL
    & \makecell[c]{
        max\_prompt\_length=8192 \\
        max\_response\_length=4096 \\
        train\_batch\_size=64 \\
    }

    & \makecell[c]{
        use\_kl\_loss=True \\
        kl\_loss\_coef=0.05
    }
    
    & same as above & same as above    \\ %
    \hline
    DPO & \makecell[c]{LLaMA \\Factory}

    & \makecell[c]{
        cutoff\_len=4096 \\
        val\_size=0.05 \\
        template=qwen
    }
    
    & \makecell[c]{
        per\_device\_train\_batch\_size=1 \\
        gradient\_accumulation\_steps=8 \\
        per\_device\_eval\_batch\_size=1 \\
        lr\_scheduler\_type=cosine \\
        warmup\_ratio=0.1
    }
    
    & \makecell[c]{
        pref\_beta=0.1 \\
        pref\_loss=sigmoid
    }
    & \makecell[c]{
        num\_train\_epochs=2 \\
        logging\_steps=1 \\
        save\_steps=96 \\
        eval\_strategy=steps \\
    }
    \\ %newline
    
    \hline
    \multicolumn{6}{c}{\textbf{SFT}} \\
    \hline
    \textbf{task} & \textbf{framework} & \textbf{data} & \textbf{model} & \textbf{optim} & \textbf{trainer} \\
    
    \cline{1-6}

    \makecell[c]{
        Checklit Verifier \\
        Claim Extractor
    }
    & VeRL
    
    & \makecell[c]{
        train\_batch\_size=64 \\
        micro\_batch\_size\_per\_gpu=1 \\
        max\_length=12288 \\
        truncation=right
    }    
    & \makecell[c]{
        model\_dtype=fp32 \\
        cpu\_offload=False \\
        offload\_params=False \\
        gradient\_checkpointing=True
    }    
    & \makecell[c]{
        lr=1e-6 \\
        betas=[0.9, 0.95] \\
        weight\_decay=0.01 \\
          warmup\_steps\_ratio=0.1 \\
        clip\_grad=1.0 \\
        lr\_scheduler=cosine
    }

    & \makecell[c]{
        total\_epochs=3 \\
        nnodes=1 \\
        n\_gpus\_per\_node=8 \\
    } 
    \\
    \hline
    Truthfulness Reward Model & VeRL
    & \makecell[c]{
        max\_length=4096
    } 
    & same as above
    & \makecell[c]{
         lr=1e-5
    } 
    & Same as above \\

    \bottomrule
    \end{tabular}}
    \caption{A systematic summary of training hyperparameter settings to ensure reproducibility.}
    \label{tab:training_parameters}
\end{table}

\subsection{Evaluation Setups}\label{sec:appendix_evaluation}
\subsubsection{Setups}
In this study, we adopt corresponding metrics for different evaluation benchmarks. For FActScore~\citep{min2023factscore}, we directly use the factscore metric proposed in the original paper, calculating the scores for each model based on its officially released 500 test data points. For the 250 samples from Hallulens-LongWiki~\citep{bang2025hallulens}, we likewise employ the three metrics recommended by the original paper: Recall@32, Precision, and F1@32. For the LongFact and Factory datasets, we adopt the evaluation pipeline proposed in VeriScore~\citep{song2024veriscore} and report three key metrics: Recall@64, Precision, and F1@64. Additionally, for the LongFact and Factory datasets, we introduce the Win Rate (WR) metric to provide a complementary performance perspective.

\textbf{FActScore.} The automated FActScore evaluation in the original paper's code\footnote{\url{https://github.com/shmsw25/FActScore}} typically relies on models such as ChatGPT for atomic fact decomposition and uses its provided outdated Wiki knowledge base for judgment. However, due to the high cost of ChatGPT and the outdated nature of the original Wiki knowledge base, we make adjustments to this evaluation pipeline.

To address these issues, our approach adopts a more cost-effective and timely solution. Specifically, we use the powerful open-source model Qwen2.5-72B-Instruct to replace ChatGPT for decomposing text into atomic facts. Concurrently, we build a local Wiki20250716 index to serve as the latest knowledge base. Based on the information retrieved from this knowledge base, we also use the Qwen2.5-72B-Instruct model to judge the veracity of each atomic fact. This pipeline ensures the evaluation's cost-effectiveness, data timeliness, and result reliability.

\textbf{Hallulens-LongWiki.} In evaluating on the Hallulens-LongWiki\footnote{\url{https://github.com/facebookresearch/HalluLens}} benchmark, we follow its standard factuality framework of claim extraction, retrieval, and verification. While we use the benchmark's official knowledge base for retrieval, we employ Qwen2.5-72B-Instruct for the extraction and verification steps to ensure a cost-effective and reproducible setup.

\textbf{LongFact.} For the LongFact\footnote{\url{https://github.com/google-deepmind/long-form-factuality}} benchmark, we randomly sample 250 instances from its more challenging ``Objects'' subset to form the test set. The SAFE framework proposed in LongFact is a complex, high-cost advanced evaluation framework that relies on Google Search. To enable quick and efficient evaluation, we instead use VeriScore to assess the LongFact dataset in this paper. VeriScore is an improved version of FActScore and SAFE, which optimizes the process by focusing on extracting and verifying ``verifiable claims" and introducing inter-sentence context to improve extraction quality, thereby avoiding unnecessary revision steps. In this paper, we consistently use the powerful Qwen2.5-72B-Instruct to complete all steps of the VeriScore evaluation.

\textbf{Factory.} Factory\footnote{\url{https://huggingface.co/datasets/facebook/FACTORY}} is a large-scale, human-verified, and challenging prompt set. We randomly select 250 samples from its ``Hard'' subset for evaluation. Similar to LongFact, we compute all three metrics using the VeriScore pipeline with a local Wiki knowledge and the Qwen2.5-72B-Instruct model.

To ensure the reproducibility of our experiments and the transparency of the evaluation pipeline, we have compiled a detailed summary of the key parameter configurations used throughout the process. These configurations cover every step from fact decomposition to final verification. The specific parameter settings are detailed in Table~\ref{tab:eval_parameters}.

\begin{table}[h]
    \centering
    \scalebox{0.9}{
    \begin{tabular}{c|c|c}  %c|ccc|
    \toprule
    \textbf{Claim Extraction} & \textbf{Wiki Retrieval} & \textbf{Claim Verification} \\
    \hline
    \makecell[c]{
        temperature=0.1 \\
        max\_tokens=8192
    } 
    & \makecell[c]{
        top\_k=10 \\
        chunk\_size=300 \\
        chunk\_overlap=20 \\
    } 
    &
    \makecell[c]{
        temperature=0.1 \\
        max\_tokens=8192
    } \\
    \bottomrule
    \end{tabular}}
    \caption{The key parameter configurations used in the evaluation pipeline.}
    \label{tab:eval_parameters}
\end{table}

\subsubsection{Metrics}
This section provides formal definitions of the evaluation metrics used in our study to assess the long-form factuality of model responses. The reported values for Precision, Recall@K, and F1@K in our experiments are the average of each metric calculated for every individual response in the test set.

\textbf{Precision.} Precision measures the factual accuracy of a generated response. It is defined as the proportion of individual facts within the response that are verifiable as supported against an external knowledge source (e.g., web search results). Let $S(y)$ be the number of supported facts in a response $y$, and $N(y)$ be the number of not-supported facts. Precision for a single response is calculated as:
\begin{equation}
    \text{Precision}(y) = \frac{S(y)}{S(y) + N(y)}
\end{equation}
A higher precision indicates that the responses contain fewer factual errors on average.

\textbf{Recall@K.} Recall evaluates the comprehensiveness of a response. In open-domain long-text generation, defining the complete set of expected facts is infeasible. Following prior work~\citep{wei2024long}, we adopt a parametric recall metric. Let $K$ be a hyperparameter representing a user's desired number of supported facts for a high-quality response. Recall@K for a single response is calculated as:
\begin{equation}
    \text{Recall@K}(y) = \min\left( \frac{S(y)}{K}, 1 \right)
\end{equation}
This metric measures whether a response provides an adequate amount of verifiable information, up to a specified limit $K$.

\textbf{F1@K.} To balance the trade-off between factual precision and informational comprehensiveness (Recall@K), we use their harmonic mean to compute the F1@K score for a single response:
\begin{equation}
    F_1@K(y) = 
\begin{cases} 
\frac{2 \cdot \text{Precision}(y) \cdot \text{Recall@K}(y)}{\text{Precision}(y) + \text{Recall@K}(y)}, & \text{if } S(y) > 0 \\
0, & \text{if } S(y) = 0 
\end{cases}
\end{equation}
A response achieves a high F1@K score only by being both highly factual and sufficiently detailed.

\textbf{WR.} Win Rate (WR) is a comparative metric designed to evaluate the performance of a candidate model (B) relative to a baseline model (A). For each prompt in the test set, responses from both models are evaluated by a judge LLM (GPT-4.1 using Prompt~\ref{prompt:wr}~\citep{alpaca_eval}), which receives two outputs and assigns a ranking based on factuality and comprehensiveness. To control for position bias, each pairwise comparison is performed twice under reversed output ordering. In the first trial, Model A’s response is provided as output1 and Model B’s as output2; we record whether B is ranked higher than A, denoted as $\text{Win}^{(1)}$. In the second trial, the order is reversed: Model B’s response is given as output1 and Model A’s as output2, producing a result $\text{Win}^{(2)}$. The win indicator for the $i$-th instance is the average of both trials:
\begin{equation}
    \text{Win} = \frac{\text{Win}^{(1)} + \text{Win}^{(2)}}{2}
\end{equation}
The overall WR of model B over A on a test set of size $N$ is defined as:
\begin{equation}
   \text{WR}_{B \text{ vs } A} = \frac{\text{Win}}{N}
\end{equation}
A WR greater than 0.5 indicates that the candidate model (B) is preferred over the baseline (A). Unlike other metrics, WR provides a single aggregate measure of relative performance over the entire test set.

\subsection{Auxiliary Rewards}\label{sec:appendix_aux_reward}
To ensure the overall quality of generated text beyond factuality and prevent potential degradation in readability or adherence to instructions, we introduce three auxiliary rewards.

\textbf{General Reward.} As we perform RL directly from the base model—bypassing SFT—a standard KL penalty is inapplicable. To prevent the policy from deviating into low-quality outputs, we employ a General Reward $R_g(\mathcal{A}_i)$ from Skywork-Reward-V2-Llama-3.2-1B~\citep{liu2025skywork} to incentivize responses that align with human preference.

\textbf{Format Reward.} We introduce a Format Reward $R_f$ to enforce structured output in our thinking model. The output must encapsulate reasoning within \textless think\textgreater \textless /think\textgreater ~tags and the final answer within \textless answer\textgreater \textless /answer\textgreater ~tags:
\begin{equation}\label{eq:format}
    R_f(o_i) = \begin{cases}
0, & \text{if}~o_i ~\text{has valid format} \\
-1, & \text{otherwise}
\end{cases}
\end{equation}

\textbf{Length Penalty.} To ensure the conciseness and information density of long-form factual responses, and to prevent the model from generating redundant content, we introduce a Length Penalty~\citep{yu2025dapo}. Let $L_m$ denote the maximum allowed length threshold and $L_c~(L_c < L_m)$ be a predefined critical length value. The Length Penalty $R_l$ is defined piecewise as follows:
\begin{equation}\label{eq:length}
    R_l(\mathcal{A}_i) = 
\begin{cases} 
0, & |\mathcal{A}_i| \leq L_m - L_c \\
\dfrac{(L_m - L_c) - |\mathcal{A}_i|}{L_c}, & L_m - L_c < |\mathcal{A}_i| \leq L_m \\
-1, & |\mathcal{A}_i| > L_m 
\end{cases}
\end{equation}

\section{Additional Experiments}
In this section, we provide additional experimental details not covered in the main body, aiming to supplement and expand upon our findings. This content primarily includes supplementary analysis of our core experiments and other results not discussed in detail in the main text.

\subsection{Verifier Training}\label{sec:verifier_training}
\textbf{Claim Extraction Model Training.} To build an efficient and lightweight factual claim extraction model, we adopt the following training pipeline. First, we leverage the Claimify method proposed in~\citep{metropolitansky2025towards}, which has a significant advantage in generating structured factual claims. We use Qwen2.5-72B-Instruct as the data construction engine, utilizing a carefully designed prompt template to obtain verifiable claims. This process ultimately results in a large-scale, high-quality training dataset containing 10187 entries. Subsequently, we use Qwen2.5-14B-Instruct as the base model and perform supervised fine-tuning on this self-constructed dataset, yielding our final lightweight claim extraction model. This model maintains high-precision extraction capabilities while significantly reducing computational overhead, allowing it to be seamlessly integrated into subsequent online reinforcement learning loops.

\textbf{Checklist Verifier Training.} To ensure the discriminative capability and generalizability of the Checklist Verifier, we train the model using a high-quality dataset of 10139 samples accumulated from previous research. First, we use the DeepSeek-R1-0528 to generate entirely new and diverse responses for each training prompt. This step effectively enhances the model's robustness in judging texts of varying styles. Subsequently, we perform post-processing on these generated responses by removing any chain-of-thought content. This directs the model’s focus toward factual verification rather than mimicking reasoning processes. Finally, we conduct supervised fine-tuning on the processed dataset using Qwen2.5-14B-Instruct as the base model, resulting in an efficient and lightweight verifier model.

We evaluate the SFT model on two self-constructed test sets: an English set with 81 questions and a Chinese set with 147 questions. The detailed evaluation results are presented in the table~\ref{tab:checklist_verifier}.

\begin{table}[h]
    \centering
    \renewcommand{\arraystretch}{1.5} %
    \scalebox{0.9}{
    \begin{tabular}{c|c|c}  %c|ccc|
    \toprule
    & \textbf{English-81Q} & \textbf{Chinese-147Q} \\
    \hline
    epoch1 & 69/81=0.852 & 128/147=0.871 \\
    epoch2 & \textbf{73/81=0.901} & \textbf{129/147=0.878} \\
    epoch3 & 70/81=0.864 & 129/147=0.878 \\
    \bottomrule
    \end{tabular}}
    \caption{Based on the evaluation of accuracy on both the English (81Q) and Chinese (147Q) test sets, we select the checkpoint from epoch 2 as the final Checklist Verifier model.}
    \label{tab:checklist_verifier}
\end{table}

\textbf{Truthfulness Reward Model Training.} The training of the Truthfulness Reward Model constitutes a crucial component of our framework. Specifically, after verifying all claims using Qwen2.5-72B-Instruct and the Wiki20250716 knowledge base, we obtain the verification results as summarized in Table~\ref{tab:claim_analysis}, where ``SUPPORT'' and ``REFUTE'' represent positive and negative samples, respectively.

\begin{table}[h]
    \centering
    \setlength{\tabcolsep}{1.2pt}
    \renewcommand{\arraystretch}{1.5} %
    \scalebox{0.9}{
    \begin{tabular}{c|ccc|ccc|ccc}  %c|ccc|
    \toprule
    & \multicolumn{3}{c|}{\textbf{ELI5}} & \multicolumn{3}{c|}{\textbf{LongFact-Gen}} & \multicolumn{3}{c}{\textbf{LongWiki-Gen}} \\
\cline{2-10}
     & SUPPORT & REFUTE & \makecell[c]{NOT\\ENOUGH\\INFO} & SUPPORT & REFUTE & \makecell[c]{NOT\\ENOUGH\\INFO} & SUPPORT & REFUTE & \makecell[c]{NOT\\ENOUGH\\INFO} \\
    \hline
    7B  & 298822 & 14758 & 57376 & 149420 & 12854 & 46928  & 50260 & 13408 & 38990 \\
    14B & 319512 & 9220 & 44304 & 168417 & 7757 & 36744  & 63920 & 9415 & 33760 \\
    32B & 308659 & 7254 & 39929 & 163364 & 6903 & 33607  & 60035 & 9980 & 32779 \\
    \bottomrule
    \end{tabular}}
    \captionsetup{width=0.9\linewidth}
    \caption{Distribution of claim verification results based on Qwen2.5-72B-Instruct and Wiki20250716.}
    \label{tab:claim_analysis}
\end{table}

Given the significant scarcity of negative samples compared to positive ones, we perform the following sampling strategy: all negative samples are duplicated three times, while positive samples are down-sampled to achieve a positive-to-negative ratio of 2:1. Through this process and based on Prompt~\ref{prompt:p_true}, we construct training datasets suitable for models of three different scales: 7B, 14B, and 32B. We then conduct SFT on pre-trained models of these three sizes to obtain reward models capable of assessing claim truthfulness based on the model's inherent knowledge boundaries.

\begin{table}[h]
    \centering
    \setlength{\tabcolsep}{2pt}
    \renewcommand{\arraystretch}{1.2} %
    \scalebox{0.9}{
    \begin{tabular}{c|ccc}  %c|ccc|
    \toprule
    & {\textbf{Model}} & \textbf{Accuracy} & \textbf{F1} \\
    \cline{1-4}
    \multirow{3}{*}{7B (3000Q)} 
    & Qwen2.5-7B & 0.7237 & 0.7684 \\
    & Base-few-shot & 0.7813 & 0.7741 \\
    & SFT & \textbf{0.8187} & \textbf{0.8272} \\
    \hline
    \multirow{3}{*}{14B (3000Q)}
        & Qwen2.5-14B & 0.6450 & 0.6321 \\
        & Base-few-shot & 0.8096 & 0.8119 \\
        & SFT & \textbf{0.8347} & \textbf{0.8421} \\
    \hline
    \multirow{3}{*}{32B (3000Q)} 
    & Qwen2.5-32B & 0.7777 & 0.7752 \\
    & Base-few-shot & 0.8200 & 0.8210 \\
    & SFT & \textbf{0.8334} & \textbf{0.8423} \\
    
    \bottomrule
    \end{tabular}}
    \caption{Performance comparison of Truthfulness Reward Models at different scales on the test sets. The results show that models after SFT significantly outperform both the base model and the base few-shot model in terms of Accuracy and F1-score.}
    \label{tab:truth_reward_model_training}
\end{table}

After model training is completed, we perform a systematic evaluation of its performance on independently constructed test sets. For the three model sizes—7B, 14B, and 32B—we construct separate test sets, each containing 3000 cases with a positive-to-negative sample ratio of 1:1. The test results of the models are detailed in Table~\ref{tab:truth_reward_model_training}. The table shows that the models after SFT exhibit significant improvements in both Accuracy and F1-score compared to the base model and the base few-shot model.

\subsection{Main Results}\label{sec:appendix_main_exp}
\begin{figure}[t]
	\centering
	\begin{minipage}[t]{0.48\linewidth}
		\centering
		\includegraphics[width=1.0\textwidth]{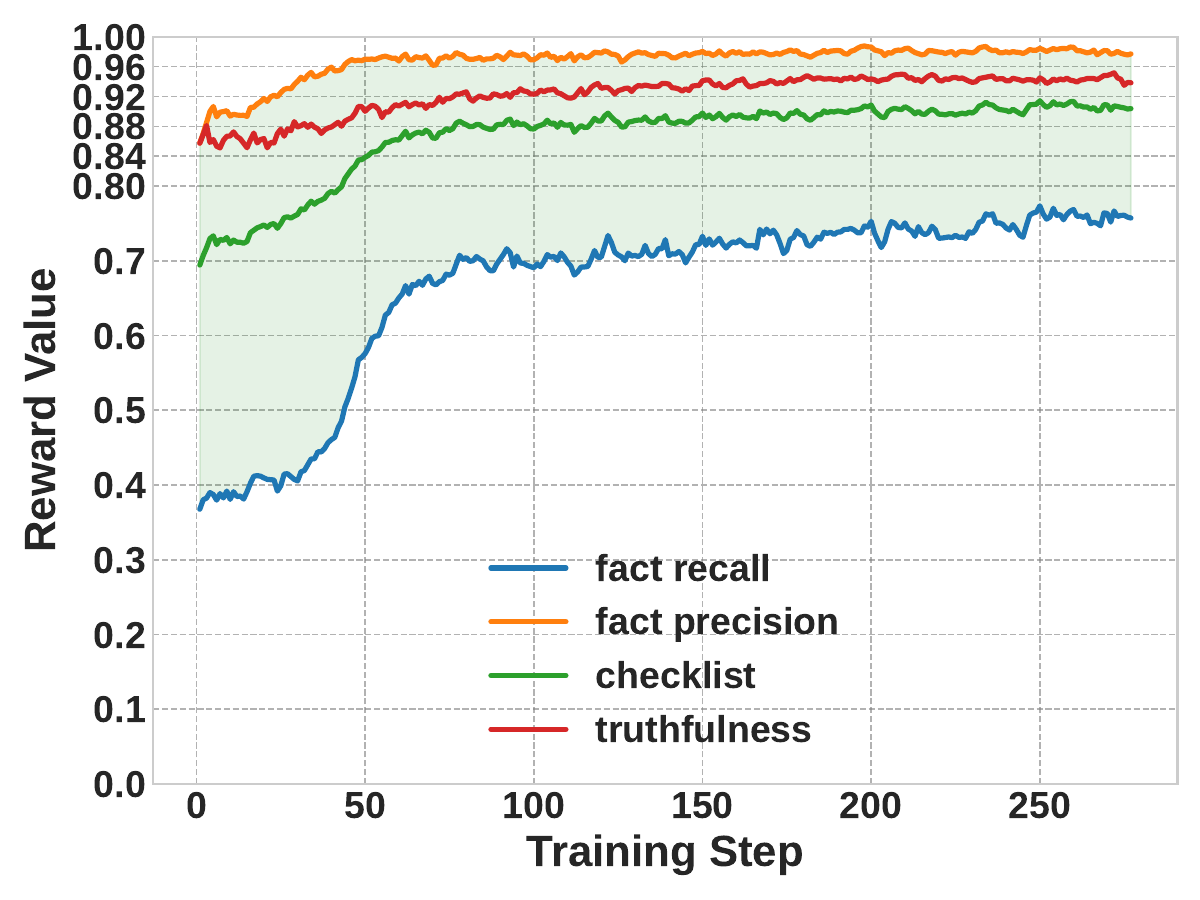}
		\begin{center}
			\footnotesize (a) Knowledge-level Consistency Rewards.
		\end{center}
	\end{minipage}
	\centering
	\begin{minipage}[t]{0.48\linewidth}
		\centering
		\includegraphics[width=1.0\textwidth]{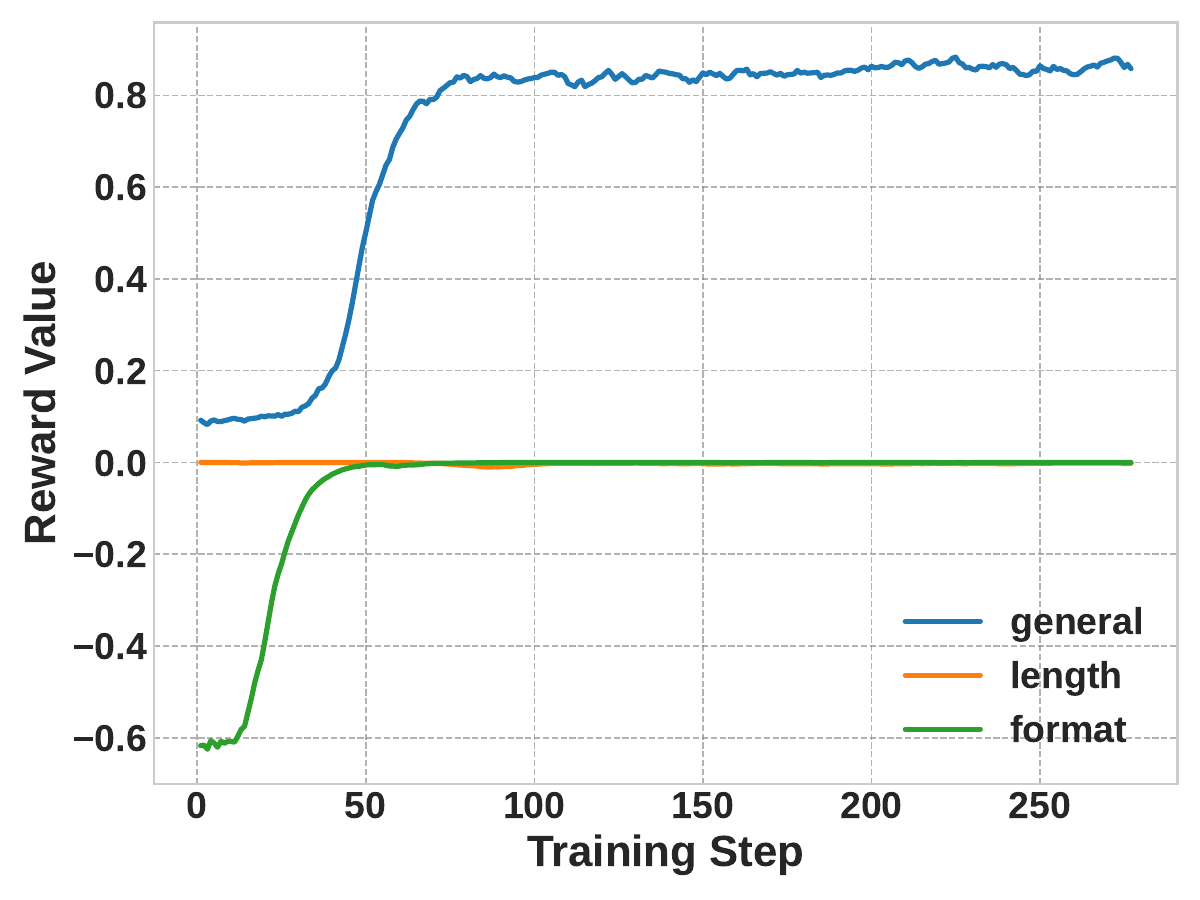}
		\begin{center}
			\footnotesize (b) Auxiliary Rewards.
		\end{center}
	\end{minipage}
    
	\centering
	\begin{minipage}[t]{0.32\linewidth}
		\centering
		\includegraphics[width=1.0\textwidth]{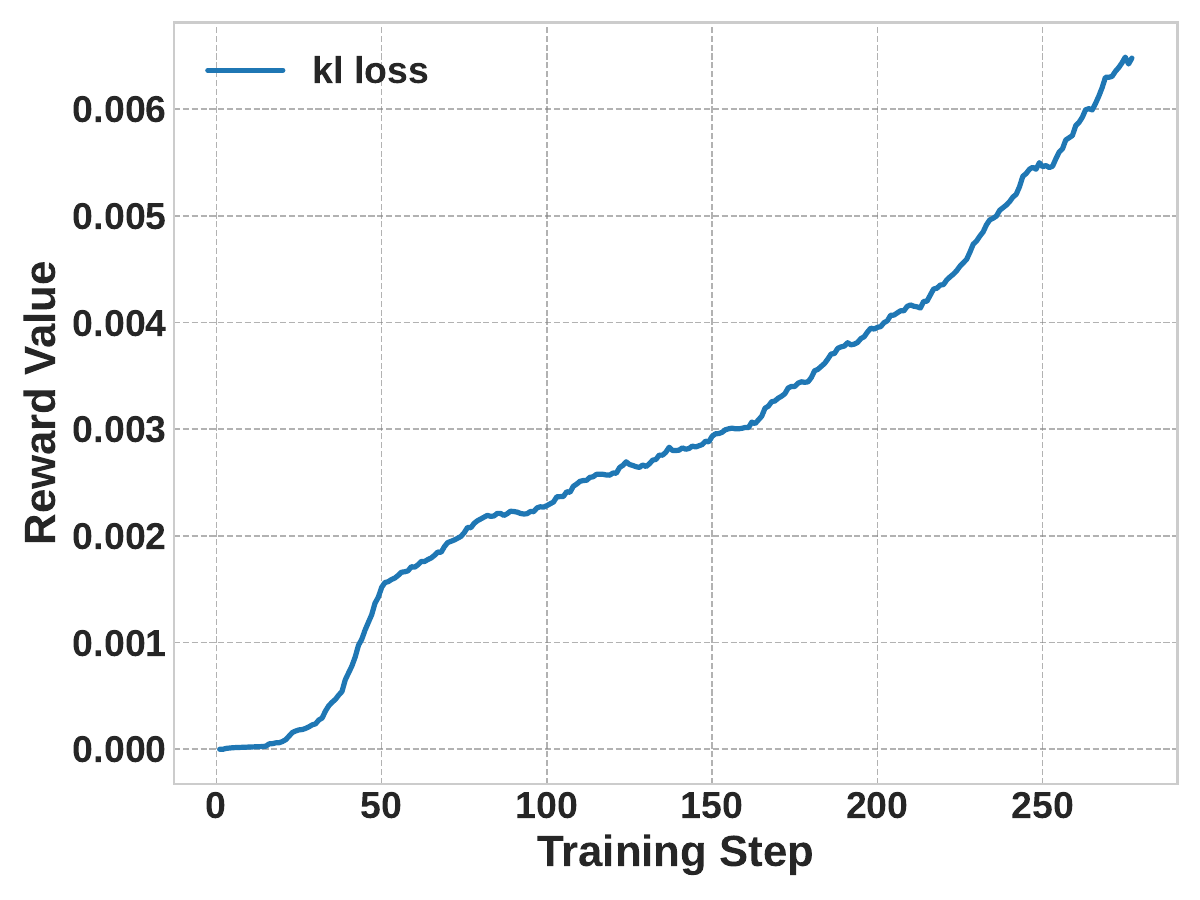}
		\begin{center}
			\footnotesize (c) KL Divergence.
		\end{center}
	\end{minipage}
	\centering
	\begin{minipage}[t]{0.32\linewidth}
		\centering
		\includegraphics[width=1.0\textwidth]{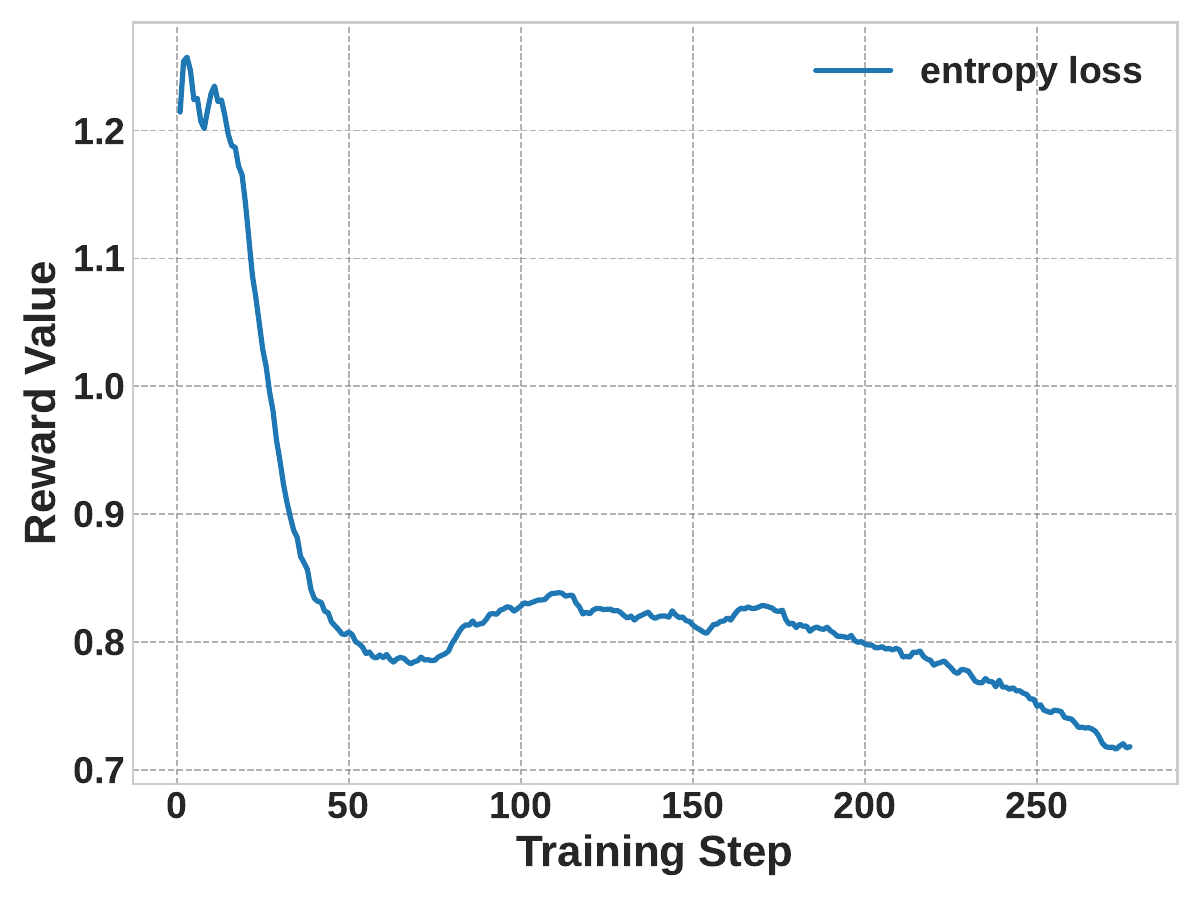}
		\begin{center}
			\footnotesize (d) Entropy Loss.
		\end{center}
	\end{minipage}
	\centering
	\begin{minipage}[t]{0.32\linewidth}
		\centering
		\includegraphics[width=1.0\textwidth]{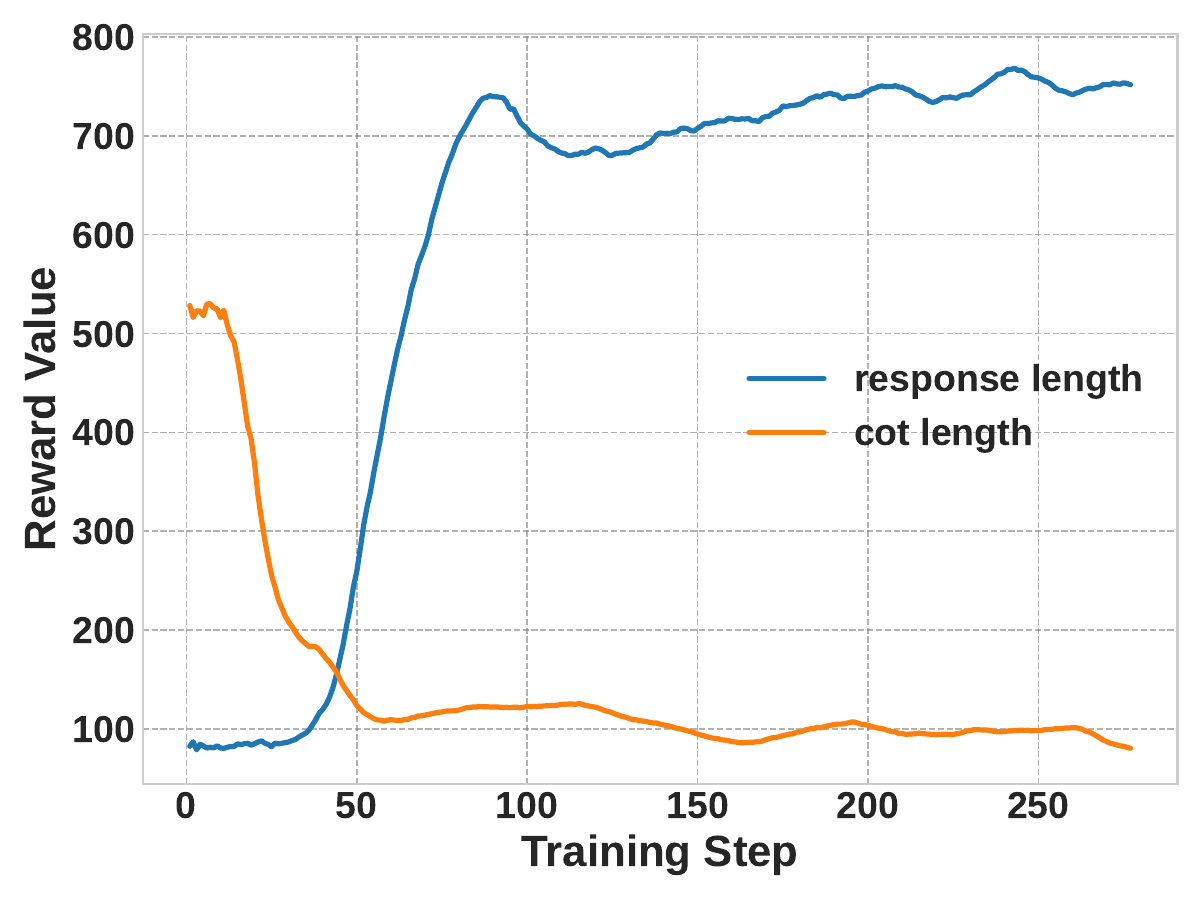}
		\begin{center}
			\footnotesize (e) Response \& CoT Length.
		\end{center}
	\end{minipage}
    \captionsetup{width=0.95\linewidth}
	\caption{Training Dynamics of KLCF-zero on Qwen2.5-14B. The figure illustrates the progression of key metrics throughout the reinforcement learning process. (a) The core knowledge-level consistency rewards, all showing significant improvement. (b) The auxiliary rewards guiding response quality and structure. (c) The KL divergence, measuring the deviation from the base model. (d) The entropy loss, reflecting the policy's exploration-exploitation balance. (e) The lengths of the generated responses and the internal reasoning chains.}\label{fig:zero_rl_14b_reward}
\end{figure}

In this section, we present a more detailed analysis of the training process and model performance through various curves that illustrate the dynamics of our KLCF framework, as summarized in Fig.~\ref{fig:zero_rl_14b_reward}.

\begin{figure}[t]
	\centering
	\begin{minipage}[t]{0.48\linewidth}
		\centering
		\includegraphics[width=1.0\textwidth]{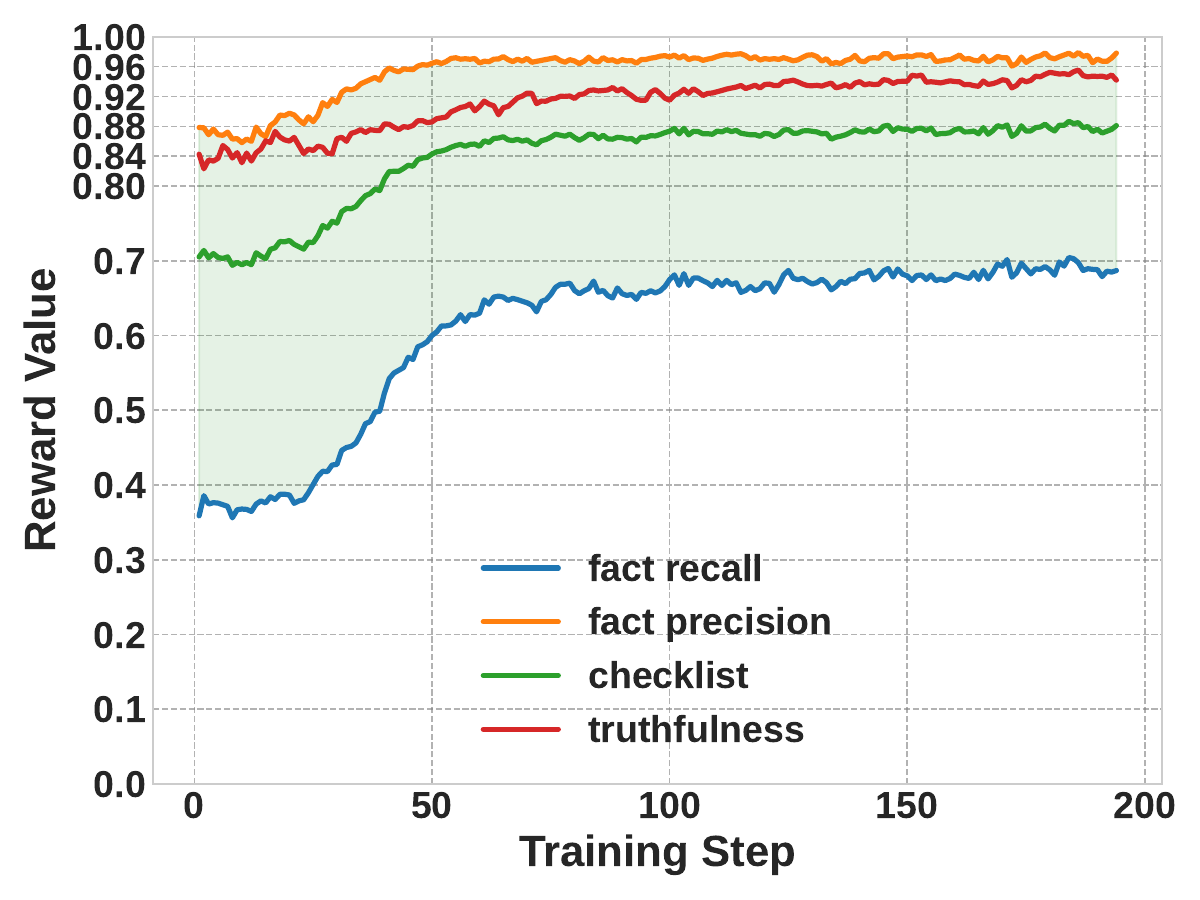}
		\begin{center}
			\footnotesize (a) Knowledge-level Consistency Rewards.
		\end{center}
	\end{minipage}
	\centering
	\begin{minipage}[t]{0.48\linewidth}
		\centering
		\includegraphics[width=1.0\textwidth]{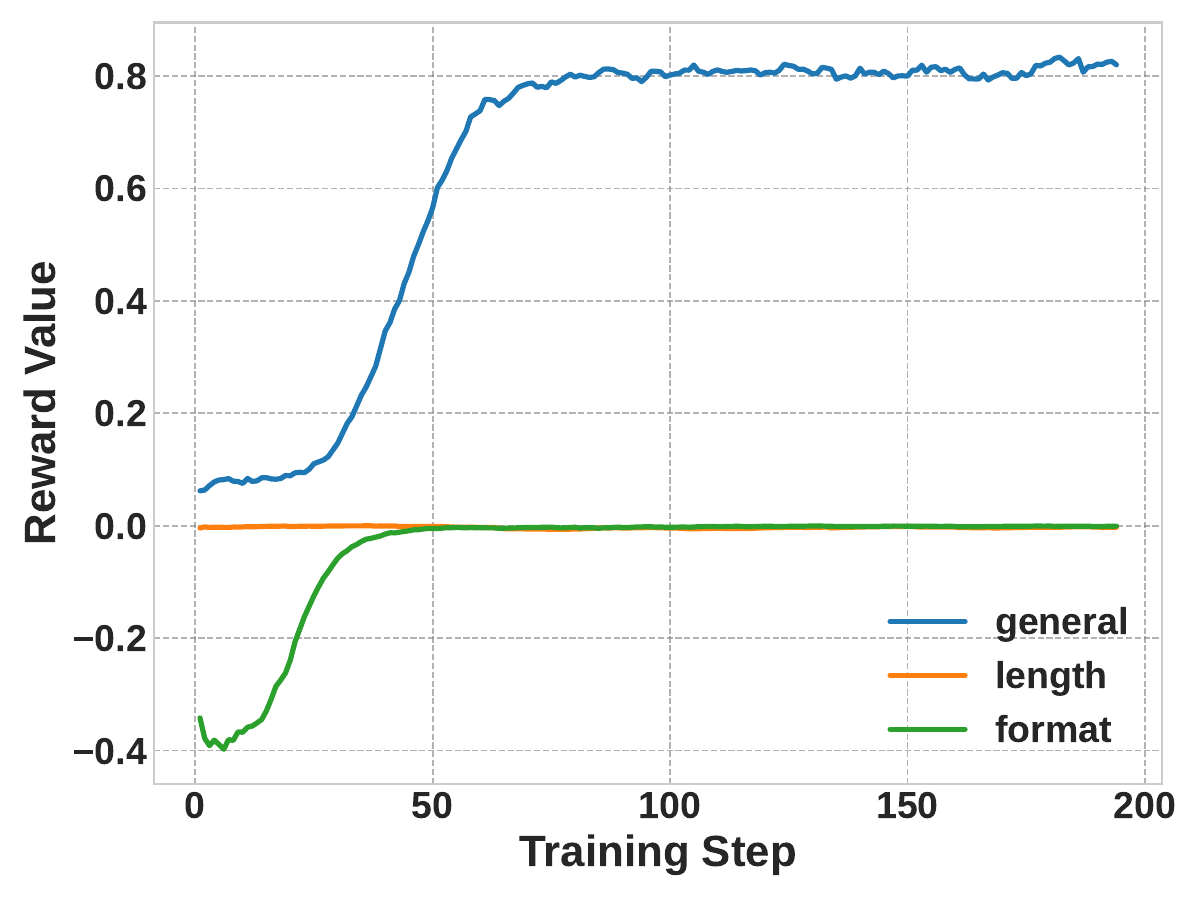}
		\begin{center}
			\footnotesize (b) Auxiliary Rewards.
		\end{center}
	\end{minipage}
    
	\centering
	\begin{minipage}[t]{0.32\linewidth}
		\centering
		\includegraphics[width=1.0\textwidth]{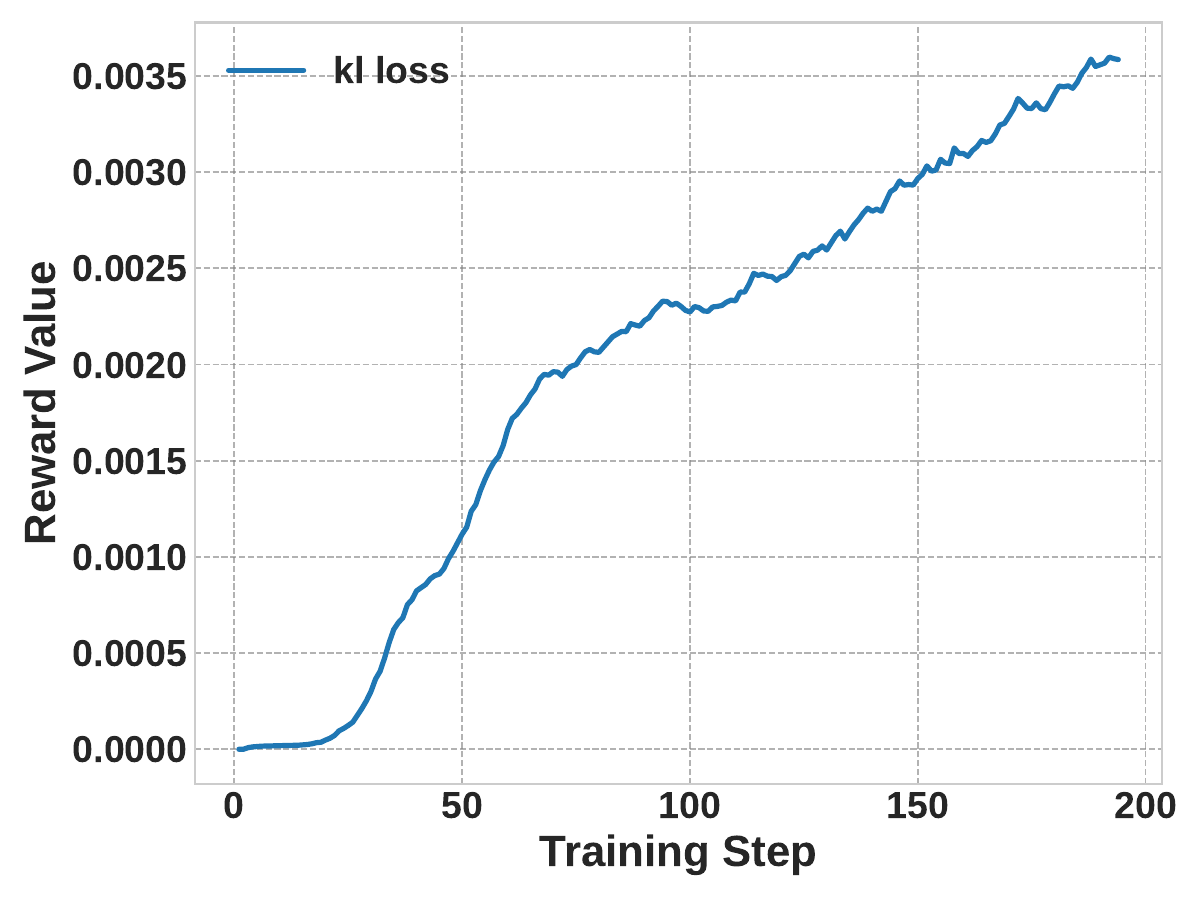}
		\begin{center}
			\footnotesize (c) KL Divergence.
		\end{center}
	\end{minipage}
	\centering
	\begin{minipage}[t]{0.32\linewidth}
		\centering
		\includegraphics[width=1.0\textwidth]{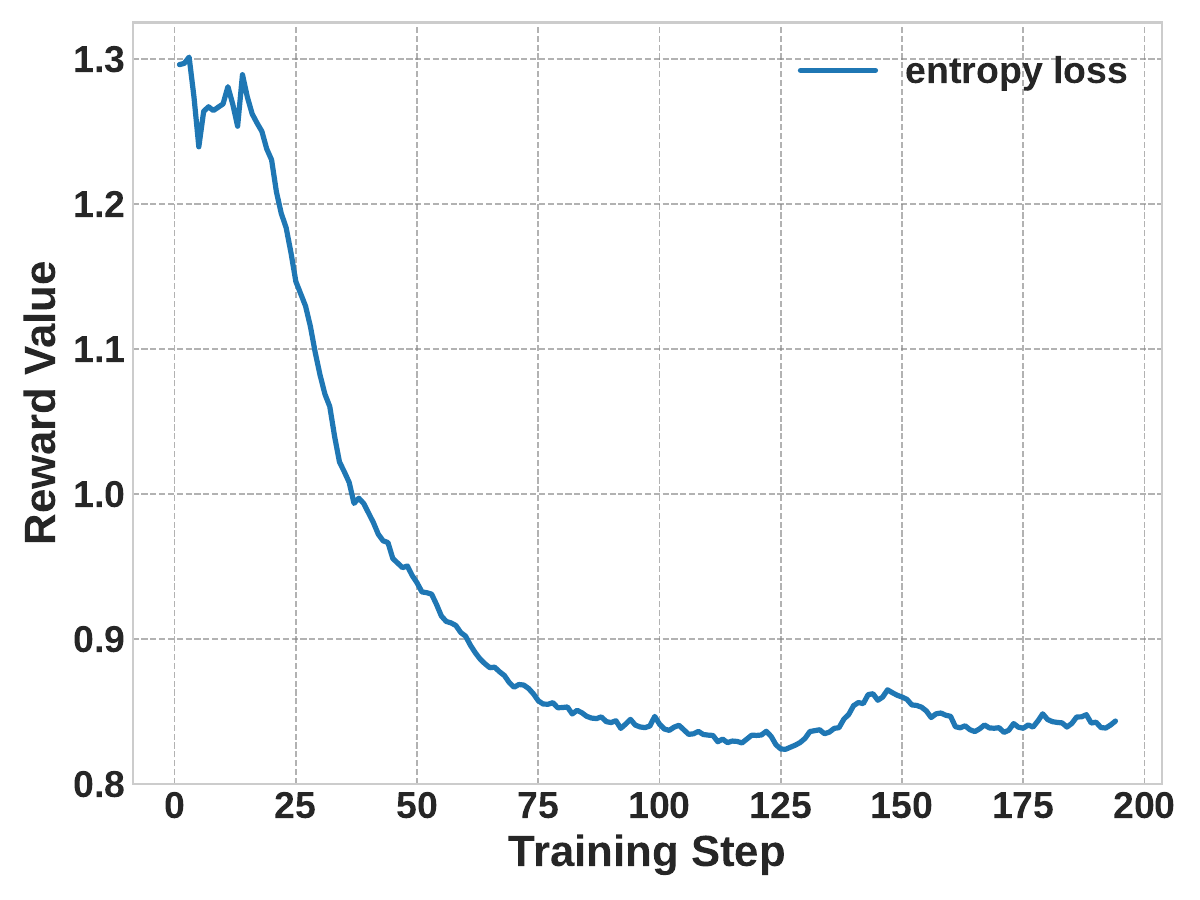}
		\begin{center}
			\footnotesize (d) Entropy Loss.
		\end{center}
	\end{minipage}
	\centering
	\begin{minipage}[t]{0.32\linewidth}
		\centering
		\includegraphics[width=1.0\textwidth]{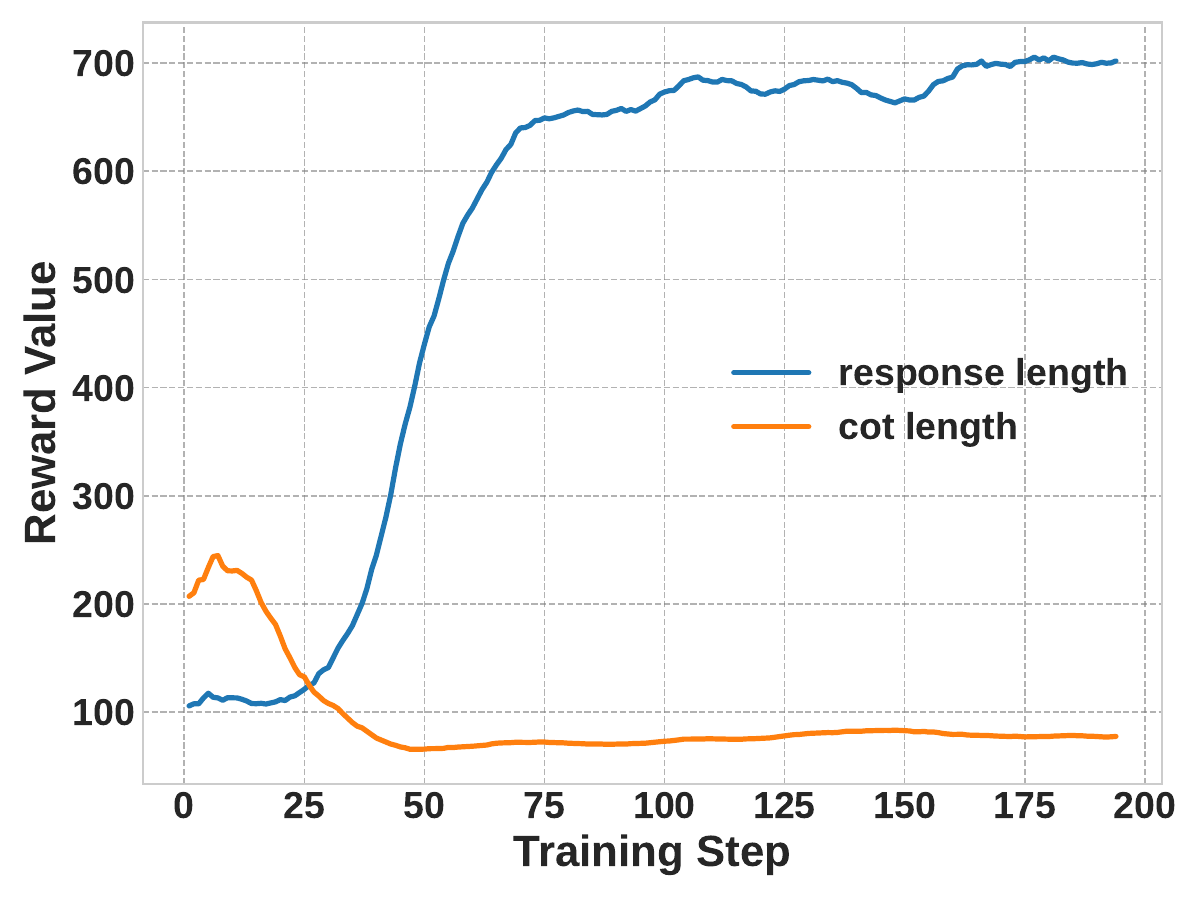}
		\begin{center}
			\footnotesize (e) Response \& CoT Length.
		\end{center}
	\end{minipage}
    \captionsetup{width=0.95\linewidth}
	\caption{Training Dynamics of KLCF-zero on Qwen2.5-7B.}\label{fig:zero_rl_7b_reward}
\end{figure}

\begin{figure}[t]
	\centering
	\begin{minipage}[t]{0.48\linewidth}
		\centering
		\includegraphics[width=1.0\textwidth]{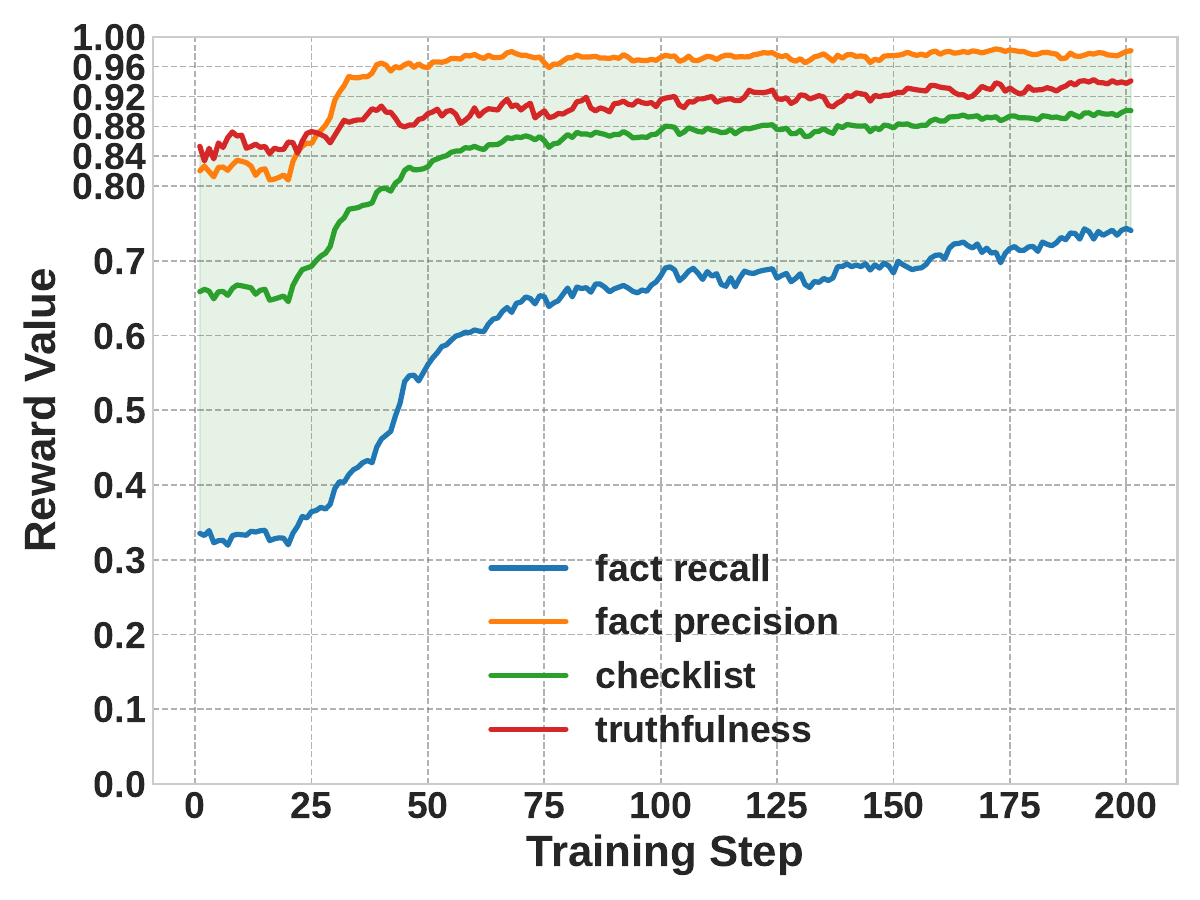}
		\begin{center}
			\footnotesize (a) Knowledge-level Consistency Rewards.
		\end{center}
	\end{minipage}
	\centering
	\begin{minipage}[t]{0.48\linewidth}
		\centering
		\includegraphics[width=1.0\textwidth]{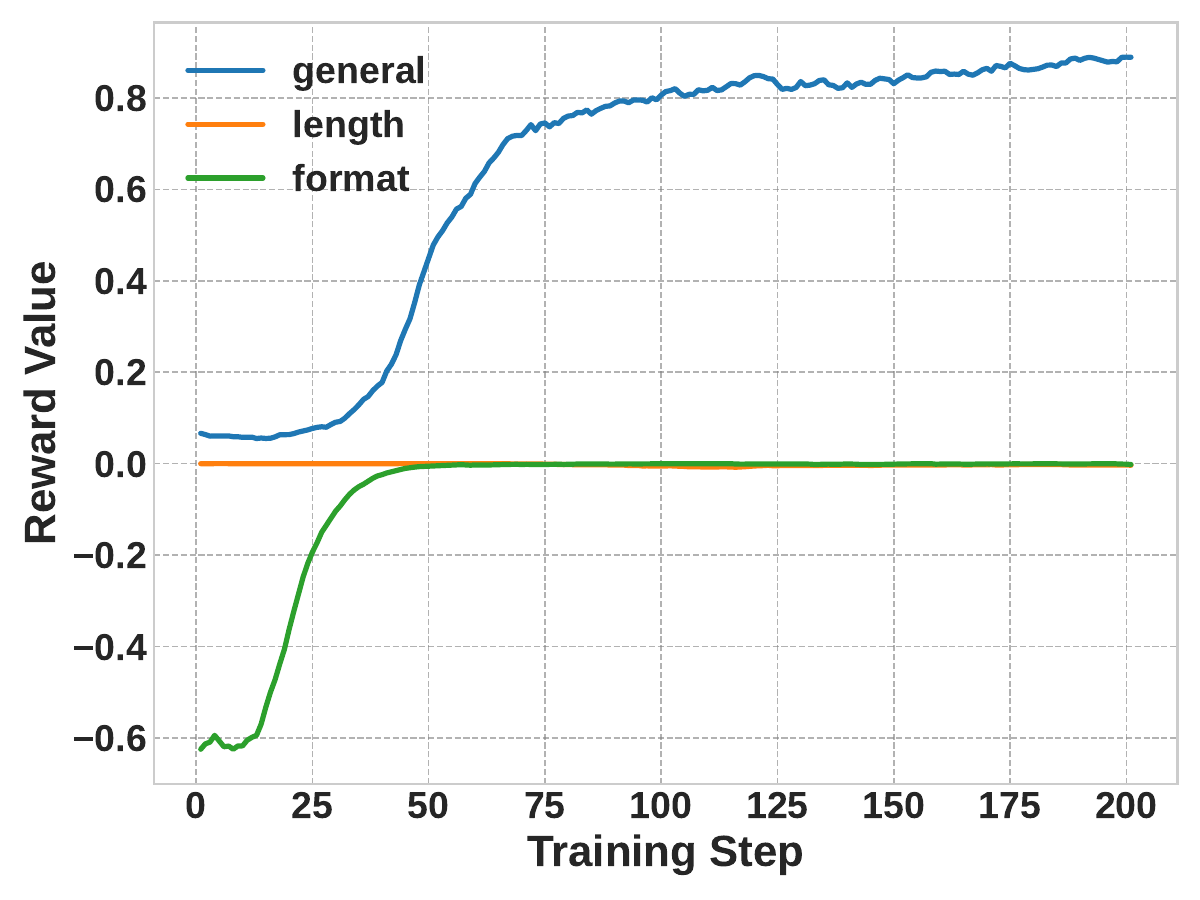}
		\begin{center}
			\footnotesize (b) Auxiliary Rewards.
		\end{center}
	\end{minipage}
    
	\centering
	\begin{minipage}[t]{0.32\linewidth}
		\centering
		\includegraphics[width=1.0\textwidth]{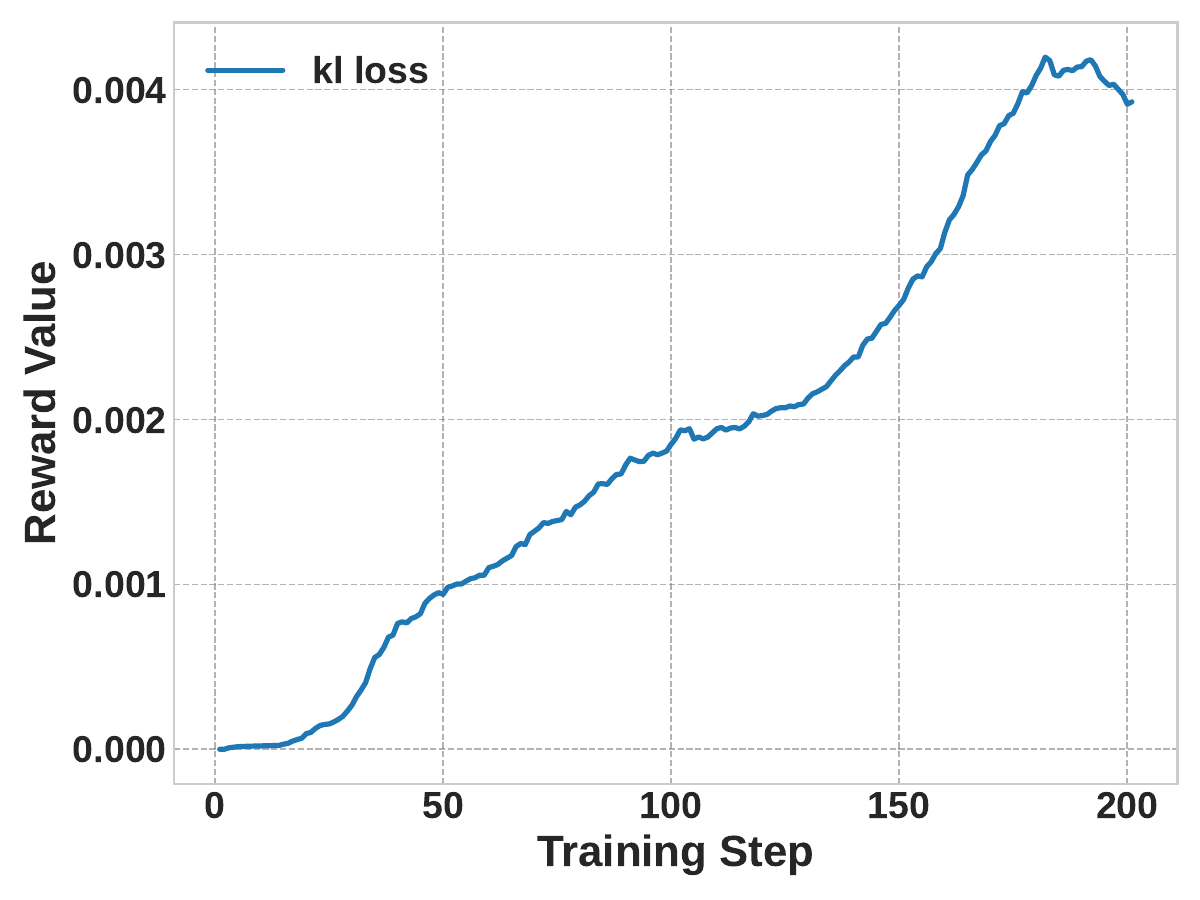}
		\begin{center}
			\footnotesize (c) KL Divergence.
		\end{center}
	\end{minipage}
	\centering
	\begin{minipage}[t]{0.32\linewidth}
		\centering
		\includegraphics[width=1.0\textwidth]{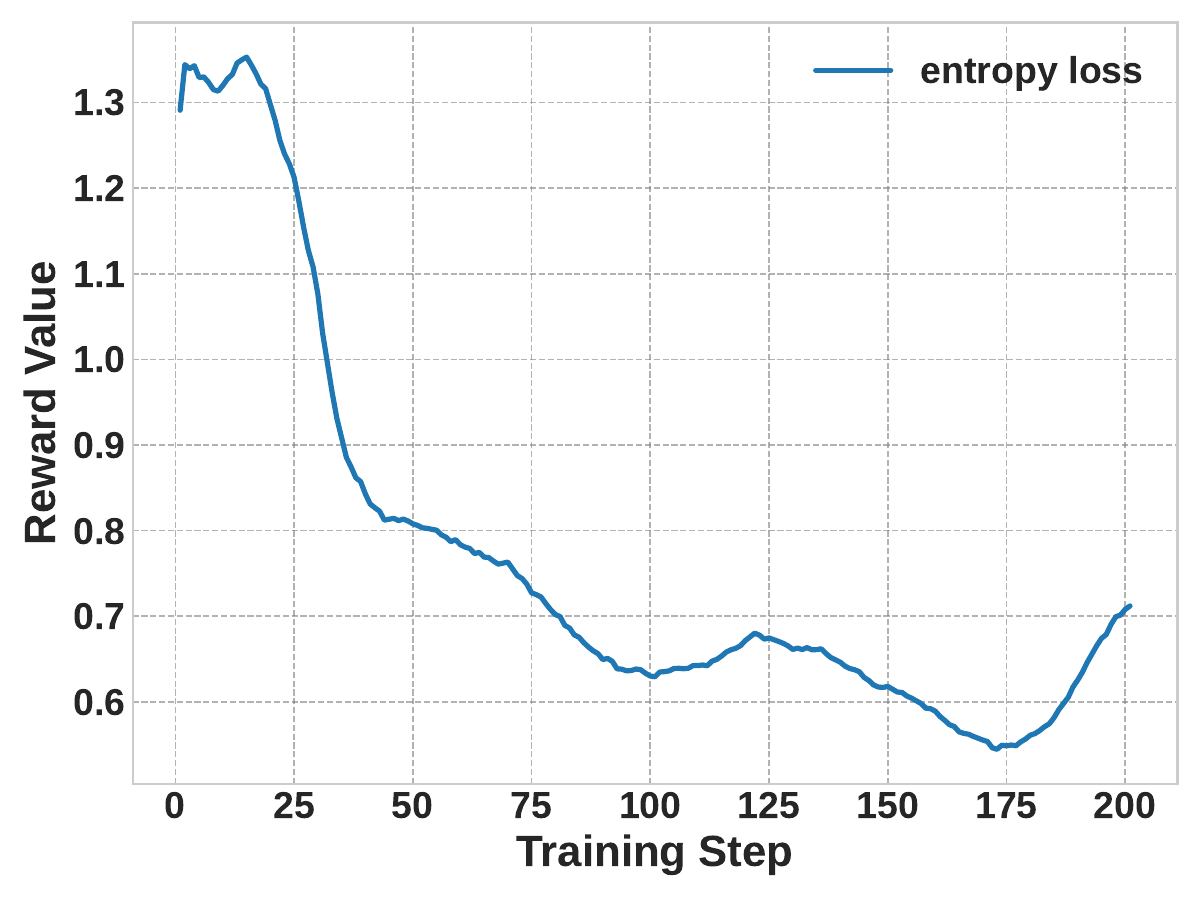}
		\begin{center}
			\footnotesize (d) Entropy Loss.
		\end{center}
	\end{minipage}
	\centering
	\begin{minipage}[t]{0.32\linewidth}
		\centering
		\includegraphics[width=1.0\textwidth]{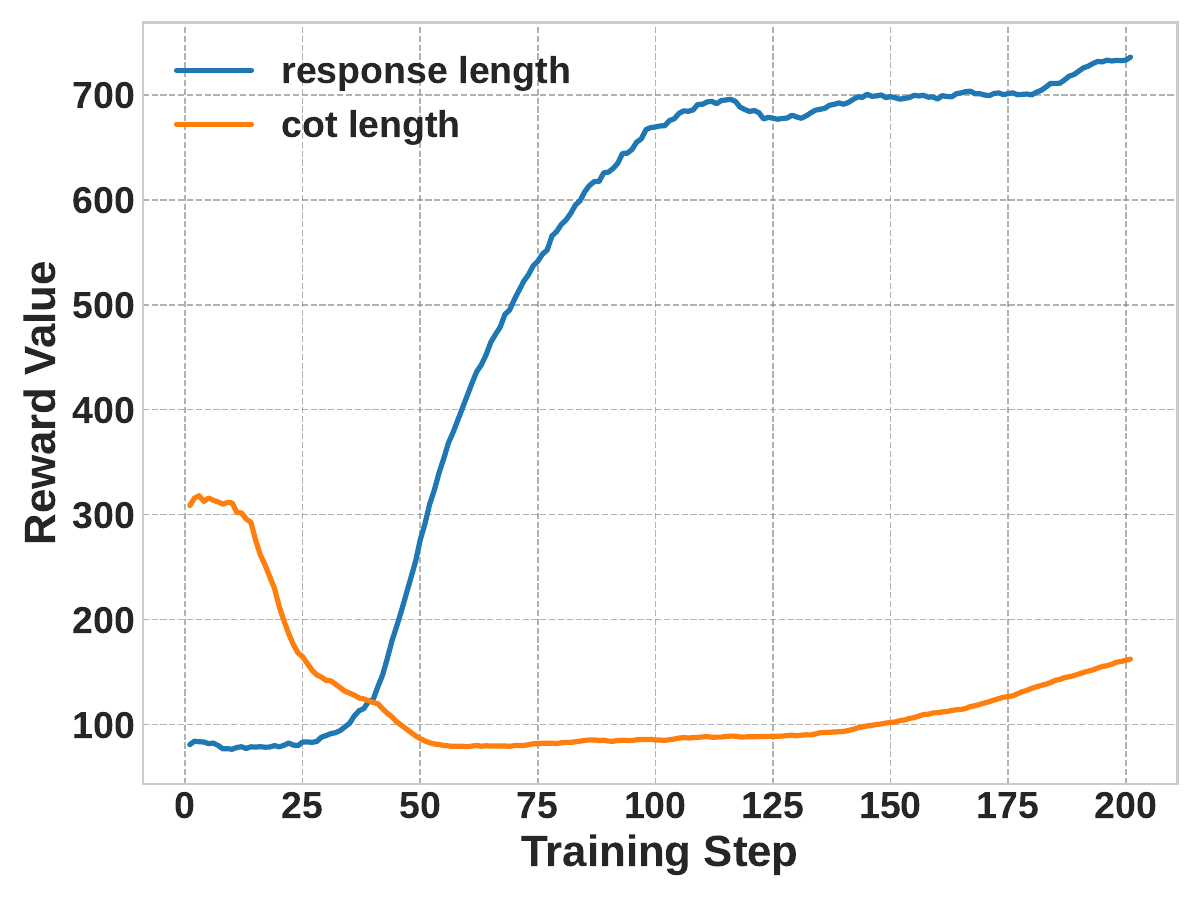}
		\begin{center}
			\footnotesize (e) Response \& CoT Length.
		\end{center}
	\end{minipage}
    \captionsetup{width=0.95\linewidth}
	\caption{Training Dynamics of KLCF-zero on Qwen2.5-32B.}\label{fig:zero_rl_32b_reward}
\end{figure}

\textbf{Knowledge-level Consistency Rewards.} All factual rewards show a significant upward trend, indicating that KLCF-zero training effectively optimizes the model’s core objectives. Among them, the \texttt{Fact Recall} reward increases markedly from approximately 0.4 to around 0.75, demonstrating the greatest improvement. This confirms that the factual coverage (comprehensiveness) of the model’s generated responses improves substantially, as the model learns to recall and output more facts within its knowledge boundaries. The \texttt{Fact Precision} reward rises further from a relatively high starting point (0.88) to near 0.98, indicating that the model maintains extremely high factual accuracy (correctness) while becoming more comprehensive. The \texttt{Truthfulness} reward remains at a very high level throughout ($0.85 \rightarrow 0.95$), reflecting the base model’s inherent strong truthfulness foundation, which our method further enhances.

\textbf{Auxiliary Rewards.} Changes in the auxiliary rewards reflect optimization of the model’s overall quality. The \texttt{General Reward} (from the Skywork-Reward model) steadily increases from 0.1 to 0.8, indicating significant improvement in the overall quality, fluency, and human preference alignment of the generated responses. The \texttt{Format Reward} converges stably from an initial penalty of approximately -0.6 to 0, indicating that the model initially made formatting errors but quickly learned and fully mastered the required output structure (\textless think\textgreater ...\textless /think\textgreater \textless answer\textgreater ...\textless /answer\textgreater), eventually incurring no further penalty. The \texttt{Length Penalty} fluctuates slightly near 0 throughout, demonstrating that the generation length remains effectively constrained within the predefined optimal range ($L_m - L_c = 850$ tokens), avoiding redundant or overly short responses.

\textbf{KL Divergence.} The KL divergence increases slowly from 0 and stabilizes at around 0.005. This slight yet consistent increase is expected and indicates that the policy model undergoes a controlled deviation from the original base model policy in order to optimize the factual rewards.

\textbf{Entropy Loss.} The entropy loss decreases rapidly in the early stages of training and then stabilizes. This trend indicates that the model quickly sharpens its policy by reducing the randomness of its choices to obtain higher rewards. The subsequent stabilization suggests that the model reaches a relatively stable optimization plateau, achieving a good balance between exploration and exploitation.

\textbf{Response Length.} The total Response Length starts at around 100 tokens, gradually increases as training progresses, and eventually stabilizes around 700 tokens. The increase in answer length is primarily driven by the recall reward, which directly incentivizes the model to cover more facts from the checklist within the <answer></answer> section. Concurrently, our designed length penalty effectively prevents answers from growing indefinitely, ensuring the conciseness and information density of the generated content.

\textbf{Regarding the shorter CoT.} With the exception of the format reward, all core factual rewards are computed solely based on the final answer and do not apply any optimization objective to the CoT itself. For general tasks like long-form QA (as opposed to complex mathematical reasoning), a longer preceding reasoning process does not necessarily lead to a better or more accurate final answer~\citep{wang2025helpsteer3}, and an ill-structured CoT can even increase the likelihood of hallucinations~\citep{cheng2025chain}. Furthermore, we observed that this phenomenon varies under different initial conditions (Base or SFT). In our experiments, when starting RL directly from the Base model, the model learned to use more concise reasoning chains to organize its thoughts during training, thereby allocating more expressive capacity to the answer section containing substantive facts. When warm-starting RL from an SFT model, the CoT length did not change significantly. We believe the effective CoT patterns for factual tasks warrant further research.

\subsection{Efficiency Analysis}\label{sec:appendix_efficiency}
\begin{table}[htbp]
\centering
\renewcommand{\arraystretch}{1.3}
\resizebox{\textwidth}{!}{%
\begin{tabular}{>{\raggedright}p{2.6cm} >{\raggedright}p{4.5cm} >{\raggedright}p{2.2cm} >{\raggedright}p{1.8cm} >{\raggedright\arraybackslash}p{4.8cm}}
\toprule
\textbf{Phase} & \textbf{Key Steps} & \textbf{Hardware Config} & \textbf{Total Time} & \textbf{Key Notes} \\
\midrule

\makecell[l]{\textbf{A. Offline Prep}\\\textbf{(One-time)}}
& \makecell[l]{1. Base Model Response Sampling\\
2. Claim Extraction\\
3. Claim Verification\\
4. Factual Checklist Construction\\
5. Truthfulness RM Training\\
6. Checklist Verifier Training}
& 4 $\times$ H100 (80GB)
& \textbf{\textasciitilde20 hours}
& All outputs (checklists, reward models) are \textbf{permanently reusable} for any subsequent RL training runs, different model scales, or even different tasks. \\
\midrule

\makecell[l]{\textbf{B. KLCF Online}\\\textbf{RL (Per Step)}}
& Policy Rollout (Response Generation)
& 16 $\times$ H100 (80GB)
& \textbf{\textasciitilde35 seconds}
& This part is \textbf{identical} to RL baselines using rewards like FActScore and represents the baseline overhead of RL training. \\
\cline{2-2} \cline{4-5}

& \textbf{KLCF Reward Computation}
&
& \textbf{\textasciitilde127 seconds}
& Uses our lightweight, local reward models; \textbf{requires no external API calls}. \\
\cline{2-2} \cline{4-5}

& Others
&
& \textbf{\textasciitilde50 seconds}
& For example, advantage estimation and parameter updates. \\
\cline{2-2} \cline{4-5}

& \textbf{KLCF Total per Step}
&
& \textbf{\textasciitilde212 seconds}
& \\
\midrule

\makecell[l]{\textbf{C. FActScore-based}\\\textbf{Online RL}\\\textbf{(Per Step)}}
& Policy Rollout (Response Generation)
& 16 $\times$ H100 (80GB)
& \textbf{\textasciitilde35 seconds}
& Same as our method. \\
\cline{2-2} \cline{4-5}

& \textbf{FActScore Reward Computation}
& (Requires External API)
& \textbf{\textasciitilde1056 seconds}
& Relies on external search and verification, subject to network latency and API rate limits. \\
\cline{2-2} \cline{4-5}

& Others
&
& \textbf{\textasciitilde50 seconds}
& \\
\cline{2-2} \cline{4-5}

& \textbf{FActScore Total per Step}
&
& \textbf{\textasciitilde1141 seconds}
& \\
\bottomrule
\end{tabular}%
}
\caption{Hardware configuration and time cost comparison across pipeline stages.}
\label{tab:pipeline-comparison}
\end{table}

This section provides additional implementation details for the efficiency analysis presented in Section~\ref{exp:efficiency}, ensuring the fairness, reproducibility, and credibility of our results.

\textbf{Data and Model Configuration.} To guarantee a broad and unbiased evaluation, we randomly select 50 prompts from the training set of the Qwen2.5-14B model to form the evaluation dataset. To completely eliminate the influence of the model's training state on the length and structure of the generated content, we use an untrained DeepSeek-R1-Distill-Qwen-14B to generate responses for all prompts. These responses serve as the unified input for all reward calculation methods.

\textbf{Our Reward Calculation Setup.} For the calculation of our proposed rewards, the scale of the reward models and verifiers remains consistent with the main experiments. Specifically, both the Checklist Verifier and the Truthfulness Reward Model are fine-tuned from the Qwen2.5-14B-Instruct and Qwen2.5-14B, respectively, ensuring that the efficiency evaluation aligns with the actual application scenario.

\textbf{Baseline Methods Calculation Setup.} To ensure a fair comparison, we standardize the internal components of the baseline methods, FActScore and VeriScore. We replace their original claim extraction and claim verification modules with the same-sized Qwen2.5-14B-Instruct. This step eliminates the potential impact of model scale differences on computational efficiency, focusing the comparison solely on the overhead introduced by the algorithmic designs.

\textbf{Deployment and Computational Environment.} We deploy all models using the vLLM inference framework on two H100 GPUs. When evaluating the parallel computing performance of all methods, we set the concurrency to 100 to effectively simulate the high-throughput reward calculation scenario common in large-scale reinforcement learning.

As shown in Table~\ref{tab:efficiency} and Table~\ref{tab:pipeline-comparison}, our KLCF reward design substantially reduces computational overhead and eliminates critical deployment bottlenecks.

\textbf{Reward computation is the dominant bottleneck in online RL training.} In every training iteration, policy rollout takes ~35 seconds, while reward calculation dominates the total step time. Our KLCF reward requires only ~127 seconds, compared to ~1056 seconds for FActScore. This means our reward computation is only \textbf{12\% of the cost of FActScore}, leading to a \textbf{5.4× faster per-iteration training} in practice.

\textbf{One‑time offline investment yields long‑term, scalable gains.} Although offline preparation takes ~20 hours (building checklists and training reward models), these assets are reusable across unlimited RL runs, model scales, and even tasks. To quantify: for a 100‑step RL run, KLCF totals ``100 × 212s = 5.89 hours'' of online time, whereas FActScore totals ``100 × 1141s = 31.69 hours''. \textbf{A single training run already saves ~25.8 hours.} For large‑scale training (thousands of steps) or hyperparameter sweeps, the savings rapidly dwarf the one‑time offline cost.

\textbf{Our design avoids real‑world deployment bottlenecks.} FActScore‑based methods rely on external search APIs, which suffer from rate limits, network latency, and instability—often becoming the slowest component in distributed RL pipelines. Moreover, in mixed‑task training (e.g., combining factual and mathematical reasoning), the entire pipeline must wait for the slowest reward signal. Our method uses only lightweight, locally deployed reward models, eliminating external API dependencies and ensuring stable, high‑throughput reinforcement learning at scale.

In summary, while we acknowledge the one‑time offline preparation cost, our method shifts the primary overhead to a reusable investment and eliminates the critical online bottleneck, achieving substantially higher overall efficiency—especially in large‑scale and mixed‑task RL settings.

\subsection{Truthfulness Reward Analysis}\label{sec:appendix_truth}
This section provides a comparative analysis of the standard Truthfulness Reward (Eq.~\ref{eq:truth}) and its variant (Eq.~\ref{eq:truth_var}), under the RL-zero training setting on Qwen2.5-14B. While both reward formulations significantly improve factuality over the base model, the variant exhibits a slight but consistent performance drop across most metrics compared to the standard version.

The observed performance gap can be primarily attributed to the inherent limitations of the factual checklist upon which the variant relies. Although the checklist is constructed from verified facts within the base model’s knowledge boundary, it is not exhaustive. Omissions or errors during the checklist construction pipeline—such as incomplete claim extraction or imperfect verification—lead to an incomplete prior. Consequently, the variant reward may incorrectly evaluate certain valid claims as “new” and subject them to the noisy estimation of the reward model, rather than leveraging them as high-confidence signals. This introduces inconsistency in the reward signal during policy optimization.

Furthermore, while the variant was designed to reduce evaluation noise by focusing only on claims outside the checklist, this very mechanism restricts its supervisory scope. The standard reward, by evaluating every claim in the response, provides a more comprehensive and stable learning signal. The variant’s narrower focus appears to slightly hinder the policy’s ability to generalize toward a balanced factuality profile, particularly in recall-oriented scenarios. Thus, despite its conceptual appeal, the variant’s dependency on a potentially imperfect checklist diminishes its effectiveness relative to the broader and more robust standard reward.

\begin{table}[t]
\centering
\setlength{\tabcolsep}{3pt}
\renewcommand{\arraystretch}{1.2} %

\scalebox{0.8}{
\begin{tabular}{c|c|ccc|cccc|cccc}
\toprule
\multirow{2}{*}{\textbf{Reward}} & \multicolumn{1}{c|}{\textbf{FActScore}} & \multicolumn{3}{c|}{\textbf{LongWiki}} & \multicolumn{4}{c|}{\textbf{LongFact}} & \multicolumn{4}{c}{\textbf{Factory}} \\
\cline{2-13}
& FS & R@32 & Prec & F1@32 & R@64 & Prec & F1@64 & WR & R@64 & Prec & F1@64 & WR \\
\hline
% \multirow{7}{*}{Qwen2.5-7B} 
Base (10-shot)     & 46.8\% & 0.661 & 0.435 & 0.511 & 0.614 & 0.685 & 0.637 & - & 0.239 & 0.302 & 0.262 & - \\
$R_{\text{truth}}$ & \textbf{61.2\%}  & \textbf{0.681} & \textbf{0.552} & \textbf{0.568} & \textbf{0.776} & \textbf{0.704} & \textbf{0.733} & \textbf{94.6\%} & \textbf{0.296} & 0.348 & \textbf{0.309} & \textbf{65.5\%} \\
$R_{\text{truth}}^{\text{variant}}$  & 60.7\% & 0.643 & 0.548 & 0.545 & 0.751 & 0.691 & 0.714 & 91.0\% & 0.284 & \textbf{0.350} & 0.299 & 56.2\% \\
\bottomrule
\end{tabular}
}

\captionsetup{width=0.9\linewidth}
\caption{Comparison between the standard truthfulness reward and its variant. Experiments were conducted by performing RL from the base Qwen2.5-14B model.}
\label{tab:truth}
\end{table}

\subsection{Case Study}\label{sec:appendix_case_study}
In this section, we conduct a qualitative study on the performance of the KLCF framework on specific samples. Specifically, we compare Qwen2.5-14B-Instruct and its KLCF-trained version on an example from the LongFact test set. The results are shown in Table~\ref{tab:case_study}.

The analysis shows that the output of the model after KLCF training is significantly longer and more detailed. This directly reflects a substantial improvement in \textbf{factual recall}, indicating that the model more actively and completely utilizes the information in its pre-trained knowledge base, rather than adopting a conservative strategy. This improvement primarily stems from our proposed Factual Checklist Reward mechanism, which incentivizes the model to cover a broader range of known facts during generation, thereby effectively mitigating over-conservatism.

More importantly, while the output content is expanded, its \textbf{factual precision} is also simultaneously improved. The output of the model without KLCF training contains some vague descriptions and information gaps, whereas the output of the KLCF model not only includes more factual details but is also described more accurately and specifically, without introducing observable new errors. This validates the effectiveness of the Dual-Fact Alignment mechanism: the Truthfulness Reward, based on the model's internal knowledge for self-assessment, guides the model to generate claims with higher confidence, thus effectively reducing the production of inaccurate or fabricated content.

In summary, this case qualitatively demonstrates that KLCF enables the model to achieve a better balance between factual recall and precision.

\begin{table}[t]
    \centering
    \scriptsize
    \begin{tabularx}{\linewidth}{>{\raggedright\arraybackslash}X}  %c|ccc|
    \toprule
    
    \textbf{Qwen2.5-14B-Instruct Response:}\\
    \hline

    Amal Clooney is a renowned international human rights lawyer and advocate based in London and New York. She is known for her work on high-profile cases involving human rights violations, freedom of expression, and other significant legal issues around the world. Amal has represented individuals and organizations in various countries, including Lebanon, Egypt, and Azerbaijan, often focusing on cases that involve press freedom and the rights of journalists.\par\medskip

\textbf{\textcolor{red}{She was born on April 18, 1978, in Beirut, Lebanon}}, and \textbf{\textcolor{red}{grew up there before moving to England at age 13}}. Amal attended \textbf{\textcolor{red}{St. Paul's Girls' School}} and later studied law at \textbf{\textcolor{red}{St Edmund Hall}}, Oxford University, and \textbf{\textcolor{red}{Middle Temple}}, where she was called to the bar in 2002. She went on to complete an LLM at \textbf{\textcolor{red}{Columbia Law School in New York City}}.\par\medskip

In addition to her legal work, Amal Clooney is also recognized for her philanthropic efforts and advocacy for various causes. She is married to the actor George Clooney, with whom she shares two children. Together, they have been involved in several charitable initiatives and campaigns aimed at promoting human rights and social justice.
    \\
    \hline
    \textbf{+ KLCF Response:} \\
    \hline

    Amal Clooney is a renowned international human rights lawyer and advocate known for her work on high-profile cases and her commitment to humanitarian issues around the world. Born on February 3, 1978, in Beirut, Lebanon, she has gained significant public attention not only due to her professional achievements but also because of her marriage to the actor George Clooney, though her career and legal work stand out as her primary contributions.\par\medskip

\#\#\# Key Aspects of Amal Clooney's Career\par\medskip

1. **Education and Early Career**: Amal Clooney earned her undergraduate law degree in Jurisprudence at St Hugh’s College, Oxford, and later obtained an LLM from New York University School of Law. Her academic background laid a strong foundation for her legal career, focusing on international law and human rights advocacy.\par\medskip

2. **Legal Expertise**: She specializes in international law, particularly in areas such as freedom of expression, press freedom, and human rights violations. She has represented clients in cases before the International Criminal Court (ICC), European Court of Human Rights, and other international tribunals.\par\medskip

3. **High-Profile Cases**: Clooney has been involved in several notable cases that have garnered global media attention. These include defending journalists and activists against charges that could be seen as politically motivated or unjust, such as the case of the Al Jazeera journalists detained in Egypt, and advocating for victims of war crimes and human rights abuses.\par\medskip

4. **Humanitarian Work**: Beyond her legal practice, she works with various organizations to promote human rights and supports initiatives aimed at protecting journalists and whistleblowers. She has been vocal about issues related to refugees and has worked on campaigns to raise awareness and improve conditions for those seeking asylum.\par\medskip

5. **Public Advocacy and Speeches**: She often speaks at international forums, conferences, and universities...\par\medskip

6. **Philanthropy and Social Causes**: Clooney is associated with multiple charities and foundations that focus on education...\par\medskip

7. **Professional Networks and Collaborations**: She collaborates with international legal bodies, non-governmental organizations...\par\medskip

8. **Personal Life and Public Image**: While her personal life, including her marriage to George Clooney...\par\medskip

\#\#\# Impact and Legacy\par\medskip

Amal Clooney's impact extends beyond individual cases; she contributes to shaping legal strategies and international norms regarding human rights and justice. Her work helps set precedents that can influence how governments and international bodies handle human rights issues and prosecute crimes against humanity.\par\medskip

Her approach to law combines rigorous legal analysis with a strategic use of media and public opinion to achieve justice for her clients and to push for broader systemic changes. This dual approach—combining traditional legal advocacy with modern communication strategies—has made her one of the most influential figures in international human rights law.\par\medskip

\#\#\# Challenges and Criticisms\par\medskip

Despite her accolades, Clooney faces criticism from some quarters regarding the effectiveness of her methods, the balance between her public persona and substantive legal outcomes, and occasional controversies surrounding the high-profile nature of her cases versus more systemic issues faced by less visible groups.\par\medskip

Overall, Amal Clooney is recognized for her dedication to advancing human rights and justice on a global scale, using her skills and platform to address complex legal and ethical challenges facing our interconnected world. Her work continues to inspire and influence legal practitioners and human rights advocates worldwide.
    \\
    
    \bottomrule
    \end{tabularx}
    \caption{Responses to the query ``\textit{Who is lawyer Amal Clooney?}'' from the Qwen2.5-14B-Instruct model and its KLCF-enhanced counterpart. Text highlighted in \textbf{\textcolor{red}{red bold}} across both responses indicates factually inaccurate claims. The KLCF-generated response is notably longer and more comprehensive, encompassing a greater number of verifiable factual details while containing fewer overall factual errors.}
    \label{tab:case_study}
\end{table}

\clearpage
\section{Prompts}\label{sec:prompts}
This section provides all prompts employed in this study. The \textcolor{red}{\textbf{red bold}} portions within the prompts indicate placeholders to be filled. 

\subsection{Data Construction}
\refstepcounter{promptcounter}
\begin{tcolorbox}[enhanced, title=Prompt for Base Model Generation, colframe=gray, colback=white, fonttitle=\bfseries]
\label{prompt:base_icl}

Please answer the following questions in a given format.
\newline
[Question]: query example 1 \newline
[Answer]: response example 1
\newline \newline
[Question]: query example 2 \newline
[Answer]: response example 2
\newline...\newline
[Question]: query example 10 \newline
[Answer]: response example 10
\newline \newline
[Question]: \textcolor{red}{\textbf{\{query\}}} \newline
[Answer]:

\end{tcolorbox}

% claim
\refstepcounter{promptcounter}
\begin{tcolorbox}[title=Prompt for Claim Extraction, colframe=gray, colback=white]
\label{prompt:claim_extract}

Below you will receive a piece of text. Your task is:\newline

1. Determine whether the text contains verifiable objective claims.\newline
2. If verifiable objective claims exist in the text, you must extract these claims from the answer (regardless of whether these claims are true).\newline
3. If the text does not contain any verifiable objective claims, return ``no verifiable objective claims''.\newline

Response format:\newline

* Claim 1\newline
* Claim 2\newline
...\newline
(or ``no verifiable objective claims")\newline

The claims you extract must adhere to the following 3 principles:\newline

1. Objectively verifiable: The claim must describe an objectively verifiable fact, not a subjective judgment, evaluation, or opinion.\newline
2. Indivisible: The objective fact described by the claim cannot be further broken down.\newline
3. Explicit meaning: Each claim must be a complete, self-contained sentence with all coreferences resolved. There should be no nouns or pronouns with unclear meaning.\newline
 
Please strictly follow the above rules to complete the following task:\newline
[Text]: \textcolor{red}{\textbf{\{response\}}}\newline
[Verifiable objective claims]:

\end{tcolorbox}

% claim verify
\refstepcounter{promptcounter}
\begin{tcolorbox}[title=Prompt for Claim Verification, colframe=gray, colback=white]
\label{prompt:claim_verify}

You will be provided with a [CLAIM] and several pieces of reference [EVIDENCE]. Your task is to determine the relationship between the facts in the [CLAIM] and the [EVIDENCE].\newline

You need to analyze each piece of [EVIDENCE] in relation to the given [CLAIM] and then provide an overall conclusion.\newline

  - **SUPPORT**: The facts in the [CLAIM] appear in the [EVIDENCE] and are consistent with the descriptions. Or, the facts in the [CLAIM] can be correctly inferred from the information in the [EVIDENCE].\newline
  - **REFUTE**: The facts in the [CLAIM] contradict the information in the [EVIDENCE]. Or, the facts in the [CLAIM] can be inferred to be false based on the information in the [EVIDENCE].\newline
  - **NOT ENOUGH INFO**: The facts in the [CLAIM] are neither supported by nor contradicted by the information in the [EVIDENCE], or a definitive conclusion cannot be drawn from the [EVIDENCE].\newline

Your response must be a dictionary with a single key-value pair, where the key is ``conclusion" and the value is the overall conclusion (**SUPPORT**, **REFUTE**, or **NOT ENOUGH INFO**). You must strictly follow the format below, returning a dictionary in JSON format. Do not return any other content.\newline

[RESPONSE FORMAT]:\newline
\verb|```|json\newline
\{\newline
\hspace*{0.5cm}``conclusion": ``The overall conclusion, return one of **SUPPORT**, **REFUTE**, or **NOT ENOUGH INFO**"\newline
\}\newline
\verb|```|\newline

Now, complete the following task.\newline
[CLAIM]: \textcolor{red}{\textbf{\{claim\}}}\newline

[EVIDENCE]:\newline
\textcolor{red}{\textbf{\{evidence\}}}\newline

Your decision:
\end{tcolorbox}

% claim importance
\refstepcounter{promptcounter}
\begin{tcolorbox}[title=Prompt for Claim Prioritization, colframe=gray, colback=white]
\label{prompt:importance}

You are given a user query and a list of candidate claims.\newline

Your task:\newline
- Retain only the claims that are essential and highly relevant to the query.\newline
- Eliminate duplicates. If two claims capture slightly different aspects, preserve both.\newline
- Each claim must convey exactly one idea, expressed as a clear, explicit, and self-contained sentence.\newline
- If conflicting claims exist, present them in the form: ``Some evidence suggests [...], while others indicate [...]".\newline
- Output format:\newline
\textbackslash boxed\{ \newline
\hspace*{0.5cm}key claim 1\newline
\hspace*{0.5cm}key claim 2\newline
\hspace*{0.5cm}key claim 3\newline
\hspace*{0.5cm}...\newline
\hspace*{0.2cm}\}\newline
- Provide output strictly within \textbackslash boxed\{\}.\newline

Output Format:\newline
\textbackslash boxed\{\newline
\hspace*{0.5cm}Final Key Claim 1\newline
\hspace*{0.5cm}Final Key Claim 2\newline
\hspace*{0.5cm}...\newline
\hspace*{0.2cm}\}\newline

Input:\newline
- Query: \textcolor{red}{\textbf{\{query\}}}\newline
- Candidate Claims:\newline
\textcolor{red}{\textbf{\{claims\}}}\newline

Output:\newline
\end{tcolorbox}

\subsection{Training}\label{sec:train_prompt}
%think prompt
\refstepcounter{promptcounter}
\begin{tcolorbox}[title=Prompt for RL-zero, colframe=gray, colback=white]
\label{prompt:rl_zero}

A conversation between User and Assistant. The user asks a question, and the Assistant solves it. The assistant first thinks about the reasoning process in the mind and then provides the user with the answer. The reasoning process and answer are enclosed within \textless think\textgreater~\textless /think\textgreater ~and \textless answer\textgreater~\textless /answer\textgreater ~tags, respectively, i.e., \textless think\textgreater ~reasoning process here \textless /think\textgreater~ \textless answer\textgreater~ answer here \textless /answer\textgreater. User: \textcolor{red}{\textbf{\{prompt\}}} Assistant: \textless think\textgreater
\end{tcolorbox}

%truthfulness rm training
\refstepcounter{promptcounter}
\begin{tcolorbox}[title=Prompt for Truthfulness Reward Model Training, colframe=gray, colback=white]
\label{prompt:p_true}

You need to evaluate the factual correctness of the following claim and output either ``True'' (the claim is entirely factually correct) or ``False'' (the claim contains errors or goes beyond your knowledge).\newline

Claim: \textcolor{red}{\textbf{\{claim\}}}\newline
Factual correctness:
\end{tcolorbox}

%checklist
\refstepcounter{promptcounter}
\begin{tcolorbox}[title=Prompt for Checklist Verification, colframe=gray, colback=white]
\label{prompt:checklist_verify}

You will receive a **Question**, a **Reply** and a **Fact List**. Your task is to check, one by one, how each fact in the list is covered in the reply and decide whether the fact is **Consistent**, **Contradictory**, or **Missing**.\newline

**Input**\newline

* [Question]: a factual question.\newline
* [Reply]: the text answer that needs to be evaluated.\newline
* [Fact List]: a list of factual points to verify.\newline

**Output**\newline

Return a list in **strict JSON format**. For every fact in the list, output a dictionary containing:\newline

* ``analysis": a brief analysis of how the reply aligns with this fact.\newline
* ``conclusion": one of ``Consistent", ``Contradictory", or ``Missing", indicating how the fact is covered in the reply.\newline

  * **``Consistent"**: the core information of the fact appears in the reply and matches the description.\newline
  * **``Contradictory"**: some information in the fact conflicts with information in the reply; from this fact you can infer that part of the reply is incorrect.\newline
  * **``Missing"**: the fact is neither fully implied nor contradicted by the reply; the reply does not mention the fact or omits some core information.\newline

Now, using the **Question**, **Reply**, and **Fact List** given below, analyze each fact and output the results strictly in the required format:\newline

[Question]: \textcolor{red}{\textbf{\{query\}}}\newline
[Reply]: \textcolor{red}{\textbf{\{response\}}}\newline
[Fact List]:\newline
\textcolor{red}{\textbf{\{guidelines\}}}\newline

[Output]:
\end{tcolorbox}

\subsection{Evaluation}\label{sec:eval_prompt}

% WR
\refstepcounter{promptcounter}
\begin{tcolorbox}[title=Prompt for Calculating Win Rate, colframe=gray, colback=white]
\label{prompt:wr}
\texttt{<|im\_start|>}system\newline
You are a helpful assistant, that ranks models by the quality of their answers.\newline
\texttt{<|im\_end|>}\newline
\texttt{<|im\_start|>}user\newline
I want you to create a leaderboard of different of large-language models. To do so, I will give you the instructions (prompts) given to the models, and the responses of two models. Please rank the models based on which responses would be preferred by humans. All inputs and outputs should be python dictionaries.\newline

Here is the prompt:\newline
\{\newline
\hspace*{0.5cm}``instruction": \texttt{"""\textbf{\textcolor{red}{\{instruction\}}}"""}\newline
\}\newline

Here are the outputs of the models:\newline
[\newline
\hspace*{0.5cm}\{\newline
\hspace*{1cm}``model": ``model\_1",\newline
\hspace*{1cm}``answer": \texttt{"""\textbf{\textcolor{red}{\{output\_1\}}}"""}\newline
\hspace*{0.5cm}\},\newline
\hspace*{0.5cm}\{\newline
\hspace*{1cm}``model": "model\_2",\newline
\hspace*{1cm}``answer": \texttt{"""\textbf{\textcolor{red}{\{output\_2\}}}"""}\newline
\hspace*{0.5cm}\}\newline
]\newline

Now please rank the models by the quality of their answers, so that the model with rank 1 has the best output. Then return a list of the model names and ranks, i.e., produce the following output:\newline
[\newline
\hspace*{0.5cm}\{`model': \texttt{<model-name>}, `rank': \texttt{<model-rank>}\},\newline
\hspace*{0.5cm}\{`model': \texttt{<model-name>}, `rank': \texttt{<model-rank>}\}\newline
]\newline

Your response must be a valid Python dictionary and should contain nothing else because we will directly execute it in Python. Please provide the ranking that the majority of humans would give.\newline
\texttt{<|im\_end|>}
\end{tcolorbox}

%% file: klcf.bib
@article{wei2024long,
  title={Long-form factuality in large language models},
  author={Wei, Jerry and Yang, Chengrun and Song, Xinying and Lu, Yifeng and Hu, Nathan and Huang, Jie and Tran, Dustin and Peng, Daiyi and Liu, Ruibo and Huang, Da and others},
  journal={Advances in Neural Information Processing Systems},
  volume={37},
  pages={80756--80827},
  year={2024}
}

@article{cotra2021ai,
  title={Why AI alignment could be hard with modern deep learning},
  author={Cotra, Ajeya},
  journal={Cold Takes},
  year={2021},
  url = "https://www.cold-takes.com/why-ai-alignment-could-be-hard-with-modern-deep-learning/",
}

@article{sharma2023towards,
  title={Towards understanding sycophancy in language models},
  author={Sharma, Mrinank and Tong, Meg and Korbak, Tomasz and Duvenaud, David and Askell, Amanda and Bowman, Samuel R and Cheng, Newton and Durmus, Esin and Hatfield-Dodds, Zac and Johnston, Scott R and others},
  journal={arXiv preprint arXiv:2310.13548},
  year={2023}
}

@article{bang2025hallulens,
  title={Hallulens: Llm hallucination benchmark},
  author={Bang, Yejin and Ji, Ziwei and Schelten, Alan and Hartshorn, Anthony and Fowler, Tara and Zhang, Cheng and Cancedda, Nicola and Fung, Pascale},
  journal={arXiv preprint arXiv:2504.17550},
  year={2025}
}

@article{fan2019eli5,
  title={ELI5: Long form question answering},
  author={Fan, Angela and Jernite, Yacine and Perez, Ethan and Grangier, David and Weston, Jason and Auli, Michael},
  journal={arXiv preprint arXiv:1907.09190},
  year={2019}
}

@misc{GoodWiki,
  title = {GoodWiki Dataset},
  author = {Choi, Euirim},
  howpublished = {\url{https://www.github.com/euirim/goodwiki}},
  month = {September},
  year = {2023}
}

@article{tian2023fine,
  title={Fine-tuning language models for factuality},
  author={Tian, Katherine and Mitchell, Eric and Yao, Huaxiu and Manning, Christopher and Finn, Chelsea},
  journal={NeurIPS 2023 Workshop on Instruction Tuning and Instruction Following},
  year={2023}
}

@article{liu2025skywork,
  title={Skywork-Reward-V2: Scaling Preference Data Curation via Human-AI Synergy},
  author = {Liu, Chris Yuhao and Zeng, Liang and Xiao, Yuzhen and He, Jujie and Liu, Jiacai and Wang, Chaojie and Yan, Rui and Shen, Wei and Zhang, Fuxiang and Xu, Jiacheng and Liu, Yang and Zhou, Yahui},
  journal={arXiv preprint arXiv:2507.01352},
  year={2025}
}

@article{yu2025dapo,
  title={Dapo: An open-source llm reinforcement learning system at scale},
  author={Yu, Qiying and Zhang, Zheng and Zhu, Ruofei and Yuan, Yufeng and Zuo, Xiaochen and Yue, Yu and Dai, Weinan and Fan, Tiantian and Liu, Gaohong and Liu, Lingjun and others},
  journal={arXiv preprint arXiv:2503.14476},
  year={2025}
}

@article{shao2024deepseekmath,
  title={Deepseekmath: Pushing the limits of mathematical reasoning in open language models},
  author={Shao, Zhihong and Wang, Peiyi and Zhu, Qihao and Xu, Runxin and Song, Junxiao and Bi, Xiao and Zhang, Haowei and Zhang, Mingchuan and Li, YK and others},
  journal={arXiv preprint arXiv:2402.03300},
  year={2024}
}

@article{wei2022chain,
  title={Chain-of-thought prompting elicits reasoning in large language models},
  author={Wei, Jason and Wang, Xuezhi and Schuurmans, Dale and Bosma, Maarten and Xia, Fei and Chi, Ed and Le, Quoc V and Zhou, Denny and others},
  journal={Advances in neural information processing systems},
  volume={35},
  pages={24824--24837},
  year={2022}
}

@article{bai2022training,
  title={Training a helpful and harmless assistant with reinforcement learning from human feedback},
  author={Bai, Yuntao and Jones, Andy and Ndousse, Kamal and Askell, Amanda and Chen, Anna and DasSarma, Nova and Drain, Dawn and Fort, Stanislav and Ganguli, Deep and Henighan, Tom and others},
  journal={arXiv preprint arXiv:2204.05862},
  year={2022}
}

@article{huang2025survey,
  title={A survey on hallucination in large language models: Principles, taxonomy, challenges, and open questions},
  author={Huang, Lei and Yu, Weijiang and Ma, Weitao and Zhong, Weihong and Feng, Zhangyin and Wang, Haotian and Chen, Qianglong and Peng, Weihua and Feng, Xiaocheng and Qin, Bing and others},
  journal={ACM Transactions on Information Systems},
  volume={43},
  number={2},
  pages={1--55},
  year={2025},
  publisher={ACM New York, NY}
}

@article{wang2024factuality,
  title={Factuality of large language models in the year 2024},
  author={Wang, Yuxia and Wang, Minghan and Manzoor, Muhammad Arslan and Liu, Fei and Georgiev, Georgi and Das, Rocktim Jyoti and Nakov, Preslav},
  journal={CoRR},
  year={2024}
}

@inproceedings{li2024flexkbqa,
  title={Flexkbqa: A flexible llm-powered framework for few-shot knowledge base question answering},
  author={Li, Zhenyu and Fan, Sunqi and Gu, Yu and Li, Xiuxing and Duan, Zhichao and Dong, Bowen and Liu, Ning and Wang, Jianyong},
  booktitle={Proceedings of the AAAI conference on artificial intelligence},
  volume={38},
  number={17},
  pages={18608--18616},
  year={2024}
}

@article{xu2024generate,
  title={Generate-on-graph: Treat llm as both agent and kg in incomplete knowledge graph question answering},
  author={Xu, Yao and He, Shizhu and Chen, Jiabei and Wang, Zihao and Song, Yangqiu and Tong, Hanghang and Liu, Guang and Liu, Kang and Zhao, Jun},
  journal={arXiv preprint arXiv:2404.14741},
  year={2024}
}

@article{zhang2024comprehensive,
  title={A comprehensive survey on process-oriented automatic text summarization with exploration of llm-based methods},
  author={Zhang, Yang and Jin, Hanlei and Meng, Dan and Wang, Jun and Tan, Jinghua},
  journal={arXiv preprint arXiv:2403.02901},
  year={2024}
}

@inproceedings{gupta2025autosumm,
  title={AUTOSUMM: A Comprehensive Framework for LLM-Based Conversation Summarization},
  author={Gupta, Abhinav and Singh, Devendra and Cowan, Greig A and Kadhiresan, N and Srivastava, Siddharth and Sriraja, Yagneswaran and Mantri, Yoages Kumar},
  booktitle={Proceedings of the 63rd Annual Meeting of the Association for Computational Linguistics (Volume 6: Industry Track)},
  pages={500--509},
  year={2025}
}

@inproceedings{zhang2025ratt,
  title={Ratt: A thought structure for coherent and correct llm reasoning},
  author={Zhang, Jinghan and Wang, Xiting and Ren, Weijieying and Jiang, Lu and Wang, Dongjie and Liu, Kunpeng},
  booktitle={Proceedings of the AAAI Conference on Artificial Intelligence},
  volume={39},
  number={25},
  pages={26733--26741},
  year={2025}
}

@article{min2023factscore,
  title={Factscore: Fine-grained atomic evaluation of factual precision in long form text generation},
  author={Min, Sewon and Krishna, Kalpesh and Lyu, Xinxi and Lewis, Mike and Yih, Wen-tau and Koh, Pang Wei and Iyyer, Mohit and Zettlemoyer, Luke and Hajishirzi, Hannaneh},
  journal={arXiv preprint arXiv:2305.14251},
  year={2023}
}

@article{chern2023factool,
  title={FacTool: Factuality Detection in Generative AI--A Tool Augmented Framework for Multi-Task and Multi-Domain Scenarios},
  author={Chern, I and Chern, Steffi and Chen, Shiqi and Yuan, Weizhe and Feng, Kehua and Zhou, Chunting and He, Junxian and Neubig, Graham and Liu, Pengfei and others},
  journal={arXiv preprint arXiv:2307.13528},
  year={2023}
}

@article{lin2024flame,
  title={Flame: Factuality-aware alignment for large language models},
  author={Lin, Sheng-Chieh and Gao, Luyu and Oguz, Barlas and Xiong, Wenhan and Lin, Jimmy and Yih, Wen-tau and Chen, Xilun},
  journal={Advances in Neural Information Processing Systems},
  volume={37},
  pages={115588--115614},
  year={2024}
}

@article{gu2025mask,
  title={Mask-dpo: Generalizable fine-grained factuality alignment of llms},
  author={Gu, Yuzhe and Zhang, Wenwei and Lyu, Chengqi and Lin, Dahua and Chen, Kai},
  journal={arXiv preprint arXiv:2503.02846},
  year={2025}
}

@article{zhang2024self,
  title={Self-alignment for factuality: Mitigating hallucinations in llms via self-evaluation},
  author={Zhang, Xiaoying and Peng, Baolin and Tian, Ye and Zhou, Jingyan and Jin, Lifeng and Song, Linfeng and Mi, Haitao and Meng, Helen},
  journal={arXiv preprint arXiv:2402.09267},
  year={2024}
}

@article{ren2025knowrl,
  title={KnowRL: Exploring Knowledgeable Reinforcement Learning for Factuality},
  author={Ren, Baochang and Qiao, Shuofei and Yu, Wenhao and Chen, Huajun and Zhang, Ningyu},
  journal={arXiv preprint arXiv:2506.19807},
  year={2025}
}

@article{ji2023survey,
  title={Survey of hallucination in natural language generation},
  author={Ji, Ziwei and Lee, Nayeon and Frieske, Rita and Yu, Tiezheng and Su, Dan and Xu, Yan and Ishii, Etsuko and Bang, Ye Jin and Madotto, Andrea and Fung, Pascale},
  journal={ACM computing surveys},
  volume={55},
  number={12},
  pages={1--38},
  year={2023},
  publisher={ACM New York, NY}
}

@inproceedings{petroni-etal-2019-language,
    title = "Language Models as Knowledge Bases?",
    author = {Petroni, Fabio  and
      Rockt{\"a}schel, Tim  and
      Riedel, Sebastian  and
      Lewis, Patrick  and
      Bakhtin, Anton  and
      Wu, Yuxiang  and
      Miller, Alexander},
    editor = "Inui, Kentaro  and
      Jiang, Jing  and
      Ng, Vincent  and
      Wan, Xiaojun",
    booktitle = "Proceedings of the 2019 Conference on Empirical Methods in Natural Language Processing and the 9th International Joint Conference on Natural Language Processing (EMNLP-IJCNLP)",
    month = nov,
    year = "2019",
    address = "Hong Kong, China",
    publisher = "Association for Computational Linguistics",
    url = "https://aclanthology.org/D19-1250/",
    doi = "10.18653/v1/D19-1250",
    pages = "2463--2473"
}

@inproceedings{yu2024reveal,
  title = {Revealing the Parametric Knowledge of Language Models: A Unified Framework for Attribution Methods},
  author = {Haeun Yu, Pepa Atanasova, Isabelle Augenstein},
  booktitle = {Proceedings of the Annual Meeting of the Association for Computational Linguistics (ACL)},
  year = {2024},
  url = {https://aclanthology.org/2024.acl-long.444.pdf},
  publisher = {Association for Computational Linguistics},
}

@article{kadavath2022language,
  title={Language models (mostly) know what they know},
  author={Kadavath, Saurav and Conerly, Tom and Askell, Amanda and Henighan, Tom and Drain, Dawn and Perez, Ethan and Schiefer, Nicholas and Hatfield-Dodds, Zac and DasSarma, Nova and Tran-Johnson, Eli and others},
  journal={arXiv preprint arXiv:2207.05221},
  year={2022}
}

@inproceedings{zhang-etal-2024-self,
    title = "Self-Alignment for Factuality: Mitigating Hallucinations in {LLM}s via Self-Evaluation",
    author = "Zhang, Xiaoying  and
      Peng, Baolin  and
      Tian, Ye  and
      Zhou, Jingyan  and
      Jin, Lifeng  and
      Song, Linfeng  and
      Mi, Haitao  and
      Meng, Helen",
    editor = "Ku, Lun-Wei  and
      Martins, Andre  and
      Srikumar, Vivek",
    booktitle = "Proceedings of the 62nd Annual Meeting of the Association for Computational Linguistics (Volume 1: Long Papers)",
    month = aug,
    year = "2024",
    address = "Bangkok, Thailand",
    publisher = "Association for Computational Linguistics"
}

@article{dhuliawala2023chain,
  title={Chain-of-verification reduces hallucination in large language models},
  author={Dhuliawala, Shehzaad and Komeili, Mojtaba and Xu, Jing and Raileanu, Roberta and Li, Xian and Celikyilmaz, Asli and Weston, Jason},
  journal={arXiv preprint arXiv:2309.11495},
  year={2023}
}

@article{guo2025deepseek,
  title={Deepseek-r1: Incentivizing reasoning capability in llms via reinforcement learning},
  author={Guo, Daya and Yang, Dejian and Zhang, Haowei and Song, Junxiao and Zhang, Ruoyu and Xu, Runxin and Zhu, Qihao and Ma, Shirong and Wang, Peiyi and Bi, Xiao and others},
  journal={arXiv preprint arXiv:2501.12948},
  year={2025}
}

@article{zhao2025learning,
  title={Learning to reason without external rewards},
  author={Zhao, Xuandong and Kang, Zhewei and Feng, Aosong and Levine, Sergey and Song, Dawn},
  journal={arXiv preprint arXiv:2505.19590},
  year={2025}
}

@article{rafailov2023direct,
  title={Direct preference optimization: Your language model is secretly a reward model},
  author={Rafailov, Rafael and Sharma, Archit and Mitchell, Eric and Manning, Christopher D and Ermon, Stefano and Finn, Chelsea},
  journal={Advances in neural information processing systems},
  volume={36},
  pages={53728--53741},
  year={2023}
}

@article{song2024veriscore,
  title={Veriscore: Evaluating the factuality of verifiable claims in long-form text generation},
  author={Song, Yixiao and Kim, Yekyung and Iyyer, Mohit},
  journal={arXiv preprint arXiv:2406.19276},
  year={2024}
}

@article{huang2024factalign,
  title={FactAlign: Long-form factuality alignment of large language models},
  author={Huang, Chao-Wei and Chen, Yun-Nung},
  journal={arXiv preprint arXiv:2410.01691},
  year={2024}
}

@article{metropolitansky2025towards,
  title={Towards Effective Extraction and Evaluation of Factual Claims},
  author={Metropolitansky, Dasha and Larson, Jonathan},
  journal={arXiv preprint arXiv:2502.10855},
  year={2025}
}

@article{chen2025factory,
  title={FACTORY: A Challenging Human-Verified Prompt Set for Long-Form Factuality},
  author={Chen, Mingda and Li, Yang and Chen, Xilun and Williams, Adina and Ghosh, Gargi and Yih, Scott},
  journal={arXiv preprint arXiv:2508.00109},
  year={2025}
}

@misc{alpaca_eval,
  author = {Xuechen Li and Tianyi Zhang and Yann Dubois and Rohan Taori and Ishaan Gulrajani and Carlos Guestrin and Percy Liang and Tatsunori B. Hashimoto },
  title = {AlpacaEval: An Automatic Evaluator of Instruction-following Models},
  year = {2023},
  month = {5},
  publisher = {GitHub},
  journal = {GitHub repository},
  howpublished = {\url{https://github.com/tatsu-lab/alpaca_eval}}
}

@inproceedings{chen2017reading,
  title={Reading {Wikipedia} to Answer Open-Domain Questions},
  author={Chen, Danqi and Fisch, Adam and Weston, Jason and Bordes, Antoine},
  booktitle={Association for Computational Linguistics (ACL)},
  year={2017}
}

@article{wang2022text,
  title={Text embeddings by weakly-supervised contrastive pre-training},
  author={Wang, Liang and Yang, Nan and Huang, Xiaolong and Jiao, Binxing and Yang, Linjun and Jiang, Daxin and Majumder, Rangan and Wei, Furu},
  journal={arXiv preprint arXiv:2212.03533},
  year={2022}
}

@article{luo2025empirical,
  title={An empirical study of catastrophic forgetting in large language models during continual fine-tuning},
  author={Luo, Yun and Yang, Zhen and Meng, Fandong and Li, Yafu and Zhou, Jie and Zhang, Yue},
  journal={IEEE Transactions on Audio, Speech and Language Processing},
  year={2025},
  publisher={IEEE}
}

@inproceedings{fu-etal-2024-disperse,
    title = "Disperse-Then-Merge: Pushing the Limits of Instruction Tuning via Alignment Tax Reduction",
    author = "Fu, Tingchen  and
      Cai, Deng  and
      Liu, Lemao  and
      Shi, Shuming  and
      Yan, Rui",
    editor = "Ku, Lun-Wei  and
      Martins, Andre  and
      Srikumar, Vivek",
    booktitle = "Findings of the Association for Computational Linguistics: ACL 2024",
    month = aug,
    year = "2024",
    address = "Bangkok, Thailand",
    publisher = "Association for Computational Linguistics",
    url = "https://aclanthology.org/2024.findings-acl.175/",
    doi = "10.18653/v1/2024.findings-acl.175",
    pages = "2967--2985"
}

@article{zhang2023language,
  title={How language model hallucinations can snowball},
  author={Zhang, Muru and Press, Ofir and Merrill, William and Liu, Alisa and Smith, Noah A},
  journal={arXiv preprint arXiv:2305.13534},
  year={2023}
}

@inproceedings{wang2025helpsteer3,
  title={Helpsteer3: Human-annotated feedback and edit data to empower inference-time scaling in open-ended general-domain tasks},
  author={Wang, Zhilin and Zeng, Jiaqi and Delalleau, Olivier and Egert, Daniel and Evans, Ellie and Shin, Hoo-Chang and Soares, Felipe and Dong, Yi and Kuchaiev, Oleksii},
  booktitle={Proceedings of the 63rd Annual Meeting of the Association for Computational Linguistics (Volume 1: Long Papers)},
  pages={25640--25662},
  year={2025}
}

@article{cheng2025chain,
  title={Chain-of-thought prompting obscures hallucination cues in large language models: An empirical evaluation},
  author={Cheng, Jiahao and Su, Tiancheng and Yuan, Jia and He, Guoxiu and Liu, Jiawei and Tao, Xinqi and Xie, Jingwen and Li, Huaxia},
  journal={arXiv preprint arXiv:2506.17088},
  year={2025}
}

@article{ni2024afacta,
  title={Afacta: Assisting the annotation of factual claim detection with reliable llm annotators},
  author={Ni, Jingwei and Shi, Minjing and Stammbach, Dominik and Sachan, Mrinmaya and Ash, Elliott and Leippold, Markus},
  journal={arXiv preprint arXiv:2402.11073},
  year={2024}
}

@inproceedings{bishop2024longdocfactscore,
  title={LongDocFACTScore: Evaluating the factuality of long document abstractive summarisation},
  author={Bishop, Jennifer A and Ananiadou, Sophia and Xie, Qianqian},
  booktitle={Proceedings of the 2024 Joint International Conference on Computational Linguistics, Language Resources and Evaluation (LREC-COLING 2024)},
  pages={10777--10789},
  year={2024}
}

@inproceedings{wanner2025dndscore,
  title={Dndscore: Decontextualization and decomposition for factuality verification in long-form text generation},
  author={Wanner, Miriam and Van Durme, Benjamin and Dredze, Mark},
  booktitle={Proceedings of the 2025 Conference on Empirical Methods in Natural Language Processing},
  pages={23620--23637},
  year={2025}
}

@article{wei2025truthrl,
  title={TruthRL: Incentivizing truthful LLMs via reinforcement learning},
  author={Wei, Zhepei and Yang, Xiao and Sun, Kai and Wang, Jiaqi and Shao, Rulin and Chen, Sean and Kachuee, Mohammad and Gollapudi, Teja and Liao, Tony and Scheffer, Nicolas and others},
  journal={arXiv preprint arXiv:2509.25760},
  year={2025}
}

@article{chen2025learning,
  title={Learning to reason for factuality},
  author={Chen, Xilun and Kulikov, Ilia and Berges, Vincent-Pierre and O{\u{g}}uz, Barlas and Shao, Rulin and Ghosh, Gargi and Weston, Jason and Yih, Wen-tau},
  journal={arXiv preprint arXiv:2508.05618},
  year={2025}
}

@article{zhang2025reinforcement,
  title={Reinforcement Learning for Better Verbalized Confidence in Long-Form Generation},
  author={Zhang, Caiqi and Zhu, Xiaochen and Li, Chengzu and Collier, Nigel and Vlachos, Andreas},
  journal={arXiv preprint arXiv:2505.23912},
  year={2025}
}

@article{li2025hallucination,
  title={The Hallucination Dilemma: Factuality-Aware Reinforcement Learning for Large Reasoning Models},
  author={Li, Junyi and Ng, Hwee Tou},
  journal={arXiv preprint arXiv:2505.24630},
  year={2025}
}

@article{prabhudesai2025maximizing,
  title={Maximizing Confidence Alone Improves Reasoning},
  author={Prabhudesai, Mihir and Chen, Lili and Ippoliti, Alex and Fragkiadaki, Katerina and Liu, Hao and Pathak, Deepak},
  journal={arXiv preprint arXiv:2505.22660},
  year={2025}
}

@article{zhang2025consistent,
  title={Consistent Paths Lead to Truth: Self-Rewarding Reinforcement Learning for LLM Reasoning},
  author={Zhang, Kongcheng and Yao, Qi and Liu, Shunyu and Wang, Yingjie and Lai, Baisheng and Ye, Jieping and Song, Mingli and Tao, Dacheng},
  journal={arXiv preprint arXiv:2506.08745},
  year={2025}
}

@article{li2025jointly,
  title={Jointly reinforcing diversity and quality in language model generations},
  author={Li, Tianjian and Zhang, Yiming and Yu, Ping and Saha, Swarnadeep and Khashabi, Daniel and Weston, Jason and Lanchantin, Jack and Wang, Tianlu},
  journal={arXiv preprint arXiv:2509.02534},
  year={2025}
}

@article{shafayat2025can,
  title={Can Large Reasoning Models Self-Train?},
  author={Shafayat, Sheikh and Tajwar, Fahim and Salakhutdinov, Ruslan and Schneider, Jeff and Zanette, Andrea},
  journal={arXiv preprint arXiv:2505.21444},
  year={2025}
}

@article{agarwal2025unreasonable,
  title={The unreasonable effectiveness of entropy minimization in llm reasoning},
  author={Agarwal, Shivam and Zhang, Zimin and Yuan, Lifan and Han, Jiawei and Peng, Hao},
  journal={arXiv preprint arXiv:2505.15134},
  year={2025}
}

@article{zuo2025ttrl,
  title={Ttrl: Test-time reinforcement learning},
  author={Zuo, Yuxin and Zhang, Kaiyan and Sheng, Li and Qu, Shang and Cui, Ganqu and Zhu, Xuekai and Li, Haozhan and Zhang, Yuchen and Long, Xinwei and Hua, Ermo and others},
  journal={arXiv preprint arXiv:2504.16084},
  year={2025}
}

@article{zhang2025right,
  title={Right question is already half the answer: Fully unsupervised llm reasoning incentivization},
  author={Zhang, Qingyang and Wu, Haitao and Zhang, Changqing and Zhao, Peilin and Bian, Yatao},
  journal={arXiv preprint arXiv:2504.05812},
  year={2025}
}

@article{ouyang2022training,
  title={Training language models to follow instructions with human feedback},
  author={Ouyang, Long and Wu, Jeffrey and Jiang, Xu and Almeida, Diogo and Wainwright, Carroll and Mishkin, Pamela and Zhang, Chong and Agarwal, Sandhini and Slama, Katarina and Ray, Alex and others},
  journal={Advances in neural information processing systems},
  volume={35},
  pages={27730--27744},
  year={2022}
}

@article{huang2025safety,
  title={Safety tax: Safety alignment makes your large reasoning models less reasonable},
  author={Huang, Tiansheng and Hu, Sihao and Ilhan, Fatih and Tekin, Selim Furkan and Yahn, Zachary and Xu, Yichang and Liu, Ling},
  journal={arXiv preprint arXiv:2503.00555},
  year={2025}
}

@inproceedings{lin2024mitigating,
  title={Mitigating the alignment tax of rlhf},
  author={Lin, Yong and Lin, Hangyu and Xiong, Wei and Diao, Shizhe and Liu, Jianmeng and Zhang, Jipeng and Pan, Rui and Wang, Haoxiang and Hu, Wenbin and Zhang, Hanning and others},
  booktitle={Proceedings of the 2024 Conference on Empirical Methods in Natural Language Processing},
  pages={580--606},
  year={2024}
}

@article{Yang2024Qwen25TR,
  title={Qwen2.5 Technical Report},
  author={Qwen An Yang and Baosong Yang and Beichen Zhang and Binyuan Hui and Bo Zheng and Bowen Yu and Chengyuan Li and Dayiheng Liu and Fei Huang and Guanting Dong and Haoran Wei and Huan Lin and Jian Yang and Jianhong Tu and Jianwei Zhang and Jianxin Yang and Jiaxin Yang and Jingren Zhou and Junyang Lin and Kai Dang and Keming Lu and Keqin Bao and Kexin Yang and Le Yu and Mei Li and Mingfeng Xue and Pei Zhang and Qin Zhu and Rui Men and Runji Lin and Tianhao Li and Tingyu Xia and Xingzhang Ren and Xuancheng Ren and Yang Fan and Yang Su and Yi-Chao Zhang and Yunyang Wan and Yuqi Liu and Zeyu Cui and Zhenru Zhang and Zihan Qiu and Shanghaoran Quan and Zekun Wang},
  journal={ArXiv},
  year={2024},
  volume={abs/2412.15115},
  url={https://api.semanticscholar.org/CorpusID:274859421}
}
